\documentclass[10pt,journal]{IEEEtran}
\usepackage[utf8]{inputenc}
\usepackage[numbers,sort&compress]{natbib}
\usepackage{amsmath,amssymb,amsfonts}
\usepackage{graphicx}
\usepackage{subfigure}
\usepackage{booktabs,tabularx}
\usepackage{algorithmic}
\usepackage{textcomp,xcolor}
\usepackage{dblfloatfix}
\usepackage[hidelinks]{hyperref}
\newcommand{\hl}[1]{#1}
\title{MonitorVLM: A Vision--Language Framework for Safety Violation Detection in Mining Operations}
\author{Jiang~Wu, Sichao~Wu, Yinsong~Ma, Guangyuan~Yu, Haoyuan~Xu, Lifang~Zheng, and Jingliang~Duan%
\thanks{Jiang Wu, Sichao Wu, Guangyuan Yu, Haoyuan Xu, Lifang Zheng, and Jingliang Duan are with the School of Mechanical Engineering, University of Science and Technology Beijing, Beijing 100083, China.}%
\thanks{Yinsong Ma is with the Laboratory for Computational Sensing and Robotics, Johns Hopkins University, Baltimore, MD 21218, USA.}%
\thanks{Jiang Wu and Sichao Wu contributed equally to this work. Corresponding author: Jingliang Duan (duanjl@ustb.edu.cn).}%
}
\begin{document}
\maketitle
\begin{abstract}
\hl{Safety monitoring in mining operations remains challenging due to complex environments and dynamic worker activities. Manual inspections are labor-intensive, error-prone, and difficult to scale, motivating automated safety analysis. This study introduces MonitorVLM, a vision--language framework for offline safety violation detection in mining surveillance videos. The framework offers three contributions: (1) a unified framework that identifies unsafe worker behaviors and links them to safety rules; (2) a domain-specific training dataset of 9,255 vision--question--answer samples covering 40 selected mining safety rules; and (3) a dual-encoder Clause Filter (CF) that dynamically selects the Top-$K$ relevant rules to support scalable rule-base expansion. Under the 40-rule setting, CF reduces average inference time by 2.8 s. The Behavior Magnifier (BM) achieves absolute percentage-point gains of 3.55 in precision and 7.31 in recall. MonitorVLM achieves 93.06\% precision, 89.59\% recall, and 91.29\% F1 score, outperforming the unfine-tuned Qwen2.5-VL-72B baseline by 19.73, 28.48, and 24.62 percentage points, respectively.}
\end{abstract}
\section{Introduction}
The vast majority of industrial accidents, including those occurring in mining and construction, originate from unsafe worker behaviors, which highlights the urgent need for continuous monitoring and timely early-warning systems~\citep{tang2020human}.  {\color{black}In practice, unsafe behaviors at industrial sites, such as workers climbing over guardrails or jumping from heights \citep{khan2023risk}, touching operating machinery with bare hands \citep{liu2023occupational}, and performing elevated work without wearing safety harnesses \citep{lee2023verification}, are still primarily detected through manual inspections conducted by human supervisors.} Such inspections are not only labor-intensive and time-consuming, but also prone to human error, inconsistency, and omissions when faced with the scale and complexity of modern worksites~\citep{rey2021design}. As industrial projects grow larger and more dynamic, these limitations make purely image-based, human-driven violation detection methods increasingly inadequate for ensuring effective and reliable safety oversight.

{\color{black}Given the limitations of manual inspection, researchers have increasingly explored automated approaches for image-based safety violation detection.} Existing methods can be broadly categorized into three groups:  (1) object-detection models, which rely on bounding-box localization to identify unsafe objects or worker postures;  (2) zero-shot semantic models, which embed images and textual prompts into a shared space to match visual content with safety clauses; and  (3) vision--language models (VLMs), which jointly process multimodal inputs to generate natural-language violation descriptions aligned with regulations.  

Early automated safety monitoring efforts largely adapted object-detection networks to industrial scenarios, with a primary emphasis on personal protective equipment (PPE) detection. Fang et al.~\citep{fang2018detecting} trained {\color{black}Faster R-CNN \citep{ren2016faster}} to locate workers' heads and classify them as ``hard-hat'' or ``no-hard-hat'', enabling automated helmet compliance checks. Building on this idea, Imam et al.~\citep{imam2025integrating} extracted human keypoints to compute joint angles and incorporated relative body size (normalized by bounding-box height) as an auxiliary feature. These signals were then integrated into the {\color{black} YOLO-Pose family (v8--v11, YOLO-World \citep{maji2022yolo}, and RT-DETR \citep{zhao2024detrs}) } for end-to-end training, allowing real-time identification of hazardous worker postures. Similarly, Vukicevic et al.~\citep{vukicevic2022generic} employed the open-source HigherHRNet~\citep{cheng2020higherhrnet} to precisely localize five body regions, including head, eyes, hands, torso, and feet, and combined this with {\color{black} MobileNetV2 \citep{sandler2018mobilenetv2}} to verify whether each of 18 custom safety items was worn in the correct location. {\color{black}Zhou et al. \citep{zhou2025integrating} extracted single frames from video streams, used YOLO to detect objects, and applied Optical Character Recognition (OCR) \citep{memon2020handwritten} to read warning signs. The results from both methods were then instantiated into a custom fire prevention knowledge graph, and rule-based reasoning was employed to output violation types and corresponding emergency measures.} Although these methods deliver high accuracy in PPE violation detection, their scope remains narrow: object detectors cannot capture contextual violations, reason about sequential actions, or interpret the deeper semantics of complex site conditions~\citep{chen2025vision,s2025review}. 

{\color{black}The advent of zero-shot semantic models, which are vision--language frameworks capable of classifying or retrieving images without task-specific training samples \citep{chen2023automatic}, has introduced a new paradigm for industrial safety inspection. These models allow images and text to be embedded within a shared feature space, enabling regulatory clauses to be matched directly with visual content.} By operating at the semantic level, {\color{black}Contrastive Language--Image Pre-training (CLIP) }\citep{radford2021learning} substantially improves both the flexibility and interpretability of violation detection. Gil~\citep{gil2024zero} demonstrated this capability by providing only two prompts, ``compliant/violation,'' allowing CLIP to score entire images without additional training or explicit annotations of PPE locations, and thus identify images containing potential hazards. Building on this idea, Hsiao et al.~\citep{hsiao2024preliminary} prefix-tuned CLIP on mining-specific image--text pairs and then fed the aligned features into a lightweight captioning module that generated natural-language descriptions with keywords such as ``no helmet'' or ``materials piled dangerously.'' These captions enabled zero-shot hazard recognition and automatic incident reporting in text form. More recently, Tsai et al.~\citep{tsai2025construction} froze the CLIP backbone and introduced two small prefix vectors to better align its embeddings with industrial safety semantics. A decoder was then used to transform the frozen image features into pixel-level captions; by scanning these captions for violation-related keywords and projecting their attention heatmaps back into the image space, the system could pinpoint the exact pixels corresponding to infractions. Despite these advances, the ability of CLIP-based methods to handle complex multimodal inputs remains limited: their similarity scores become unstable with moderately complex prompts, and, because they treat each frame independently, they fail to exploit temporal cues or capture sequential worker actions.

The recent development of large-scale vision--language models (VLMs), such as GPT-4o~\citep{yan2025gpt}, Claude~\citep{anderson2025comparative}, and Qwen2.5-VL~\citep{bai2025qwen2}, has further advanced cross-modal reasoning for safety inspection tasks. Adil et al.~\citep{adil2025using} applied prompt engineering to industry-hazard images using GPT-4o and Gemini-1.5-Pro; without any task-specific training \cite{lu2025omnitester}, the models generated natural-language alerts, such as ``missing toe board on scaffold,'' that aligned strictly with on-site regulations, illustrating strong cross-site transferability. Chaudhary et al.~\citep{chaudhary2025prompt} systematically evaluated five state-of-the-art VLMs (Claude-3 Opus, GPT-4.5, GPT-4o, GPT-o3, and Gemini 2.0 Pro) on 1,150 real-world hazard images under zero-shot, few-shot, and chain-of-thought prompting, showing that {\color{black}Chain-of-Thought (CoT) }consistently improved precision across all models. Beyond zero-shot use, fine-tuning on domain-specific datasets~\citep{zhang2024vision} has been shown to further enhance anomaly detection in industrial contexts. For example, Chen et al.~\citep{chen2025tailored} applied parameter-efficient {\color{black}Low-Rank Adaptation (LoRA)} fine-tuning to Qwen2-VL-7B on 1,139 images spanning 31 hazard categories, raising zero-shot precision from 0.495 to 0.856. {\color{black}Chan et al. \citep{chan2025context} proposed a two-stage fine-tuning approach. First, the model learns construction-domain knowledge by training on construction images and descriptions collected from the internet. Then, GPT-4V \citep{yan2025gpt} and GroundingDINO \citep{liu2024grounding} are used to automatically generate violation-related question--answer pairs for task-level fine-tuning in the safety domain. This method achieves strong performance across more than a dozen categories of high-altitude operation detection.
Despite these advances, there remain two major gaps for safety-oriented VLMs in real deployments. First, naively injecting the full set of regulatory clauses into the prompt results in significant inference latency as the rule base scales to hundreds or thousands of entries, with the added drawback of diluted clause-specific attention. Second, surveillance footage in mining environments often suffers from degraded visual quality, such as long-range viewpoints, low illumination, and occlusions, which undermines fine-grained action recognition and reliable understanding of subtle safety-critical behaviors.}

To address these challenges, we propose MonitorVLM, a novel vision--language framework for industrial violation detection, with experiments conducted on surface and underground mining operations. By rethinking both training and inference, MonitorVLM is able to recognize workers' dynamic behaviors and map them directly to violated safety clauses. Our main contributions are summarized as follows:
\begin{enumerate}[1)]
    \item \hl{We propose MonitorVLM, a pioneering vision--language framework specifically designed for automated safety violation detection in offline mining surveillance videos. By identifying unsafe worker behaviors and associating them with relevant safety regulations, the framework provides an automated alternative to labor-intensive manual video inspection.}

    \item \hl{We construct a vision--question--answer dataset comprising 3,085 original training samples covering 40 high-frequency and high-risk mining safety regulations. To enhance data diversity and model robustness, we employ targeted image augmentation and auxiliary bounding-box annotations generated by an open-vocabulary detector. Combined with LoRA fine-tuning, these strategies improve precision and recall by 10.53 and 16.81 percentage points, respectively, compared with fine-tuning on the original dataset alone.}
    
    \item \hl{To accelerate inference, we design an intelligent clause filter (CF) that selects the Top-$K$ most relevant safety clauses for each video frame. The CF is implemented on top of a dual-path model that encodes both visual features and clause-text embeddings, and outputs relevance scores for dynamic clause filter. By drastically reducing the prompt length fed to the VLM, the CF achieves a 13.73\% reduction in inference time without sacrificing precision and recall.}

\end{enumerate}

Experimental results demonstrate that MonitorVLM can rapidly detect violations in video streams and precisely identify the specific clauses breached by workers. We further implement MonitorVLM as a lightweight web-based interface, allowing users to upload videos and automatically generate timestamped violation reports with a single click. A demonstration of the interface is available at {\color{blue}\href{https://drive.google.com/file/d/1Qj23DLqOToCt8VlPdW0eGLSVEGbx20kR/view?usp=sharing}{MonitorVLM}}. The remainder of this paper is organized as follows: Section~\ref{PRELIMINARIES} introduces the preliminaries, Section~\ref{Methodology} describes the proposed methodology, Section~\ref{Experiments} reports the experimental results, and Section~\ref{Conclusion} concludes the paper.

\section{PRELIMINARIES}
\label{PRELIMINARIES}
\subsection{Problem description}
\begin{figure*}[!htpb]        
  \centering
  \includegraphics[width=0.95\textwidth]{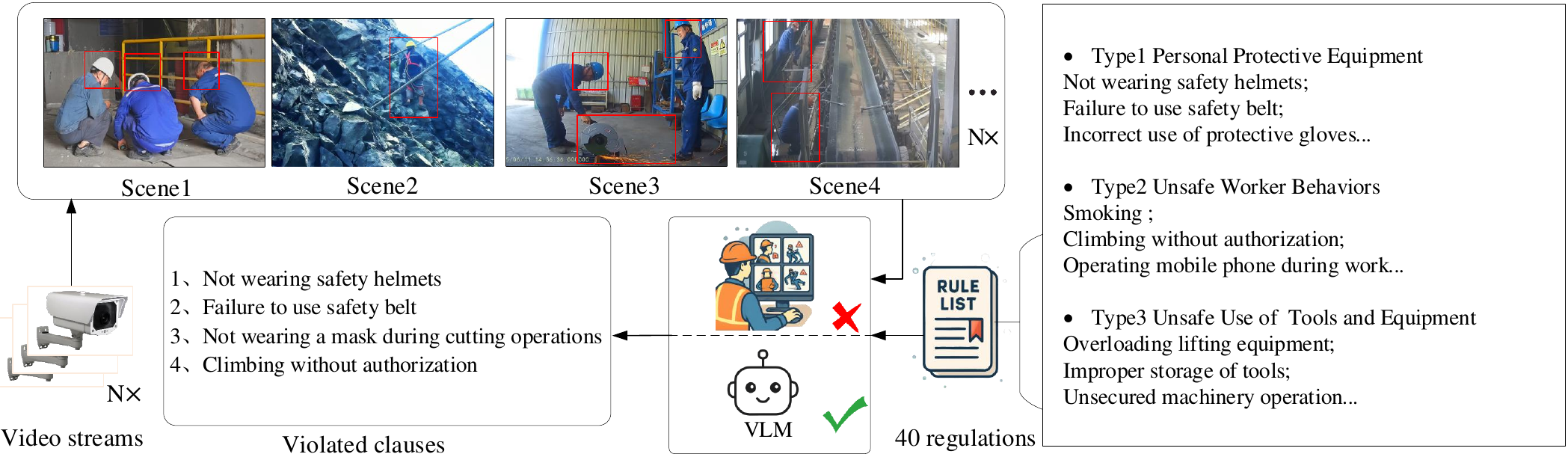}
  \caption{Illustration of industrial safety inspection tasks. The left panels show representative violation video frames, while the right panel summarizes excerpts from the 40 high-frequency regulations adopted in this study.}
  \label{fig:manual analysis}
\end{figure*}

   {\color{black}Safety violation detection is particularly critical in both surface and underground mining operations due to inherently hazardous working environments. Surface sites involve heavy machinery, blasting activities, and unstable terrain across expansive areas, whereas underground settings further impose confined spaces, limited visibility, poor ventilation, and restricted evacuation routes \citep{qiu2024study}. These concurrent operations and rapidly evolving conditions make continuous monitoring indispensable, but manual inspection alone is often insufficient at scale.} 
On a typical mining site, hundreds of types of violations may occur, each corresponding to a specific regulatory clause. As illustrated in Fig.~\ref{fig:manual analysis}, which presents four violation videos, traditional manual inspection requires experts to review video footage frame by frame, mentally align each scene with the relevant clause, and then classify the violation type. This process is not only inefficient but also prone to omissions \citep{cheng2022vision, johansen2024automated}.

\hl{In this study, we focus on 40 high-frequency mining safety clauses (the full clause list is provided in Table~}
\ref{tab:unsafe_practices}
\hl{ of the Appendix). These clauses are selected primarily based on the Chinese national mandatory standard (GB 16423-2020)~}
\citep{GB16423_2020}
\hl{, supplemented by accident investigation reports and on-site violation statistics from mining operations. The selected violations encompass different levels of detection complexity, including those identifiable from individual frames, those requiring spatial reasoning about workers and equipment, and those involving temporal or operational context. As the rule set expands to hundreds of clauses, semantic overlap and ambiguity may become more pronounced, making clear and machine-interpretable rule definitions particularly important. Therefore, we prioritize safety clauses with well-defined criteria and minimal semantic overlap to facilitate accurate model interpretation and reduce ambiguity in violation detection. Our objective is to develop an automated framework for detecting safety violations in offline surveillance videos, reducing the workload associated with manual video inspection in mining operations.}

\subsection{VLMs and fine-tuning methods}
Defined as AI systems that seamlessly integrate computer vision and natural language processing, VLMs provide a unified understanding of both visual and textual modalities \citep{danish2025comprehensive,sohail2025convlm}. Equipped with strong cross-modal reasoning capabilities, VLMs demonstrate preliminary ability to analyze violation videos, assess regulatory clauses, and generate clause-specific explanations.  

   {\color{black}Although VLMs perform well in general open-domain tasks, their effectiveness in specialized domains is often limited due to the distribution gap between pre-training data and domain-specific scenarios \citep{zhang2024vision, chen2025vision}.} Domain-specific fine-tuning is therefore indispensable to fully unlock their potential. However, due to their billions of parameters, full fine-tuning is computationally prohibitive. As one of the most widely adopted fine-tuning strategies in recent years, {\color{black}Low-Rank Adaptation (LoRA)} has emerged as a parameter-efficient alternative \citep{hu2022lora}. 
Given a pre-trained weight matrix $W_0 \in \mathbb{R}^{d \times k}$, LoRA models the weight update as:
\begin{equation}
\Delta W = \frac{\alpha}{r} BA
\end{equation}
where $A \in \mathbb{R}^{r \times k}$ and $B \in \mathbb{R}^{d \times r}$, with rank $r \ll \min(d,k)$. The forward pass is then given by:
\begin{equation}
h = W_0 x + \frac{\alpha}{r} BAx
\end{equation}
The key insight is that weight updates during fine-tuning often have low intrinsic rank $r$, enabling effective adaptation with much smaller parameter sets. This approach reduces trainable parameters to 0.1\%-1\% of the original model while maintaining performance, making it particularly suitable for adapting large models like VLMs to specialized domains such as safety inspection.

\section{Methodology}
\label{Methodology}
\begin{figure*}[!htbp]
  \centering
  \includegraphics[width=0.95\textwidth]{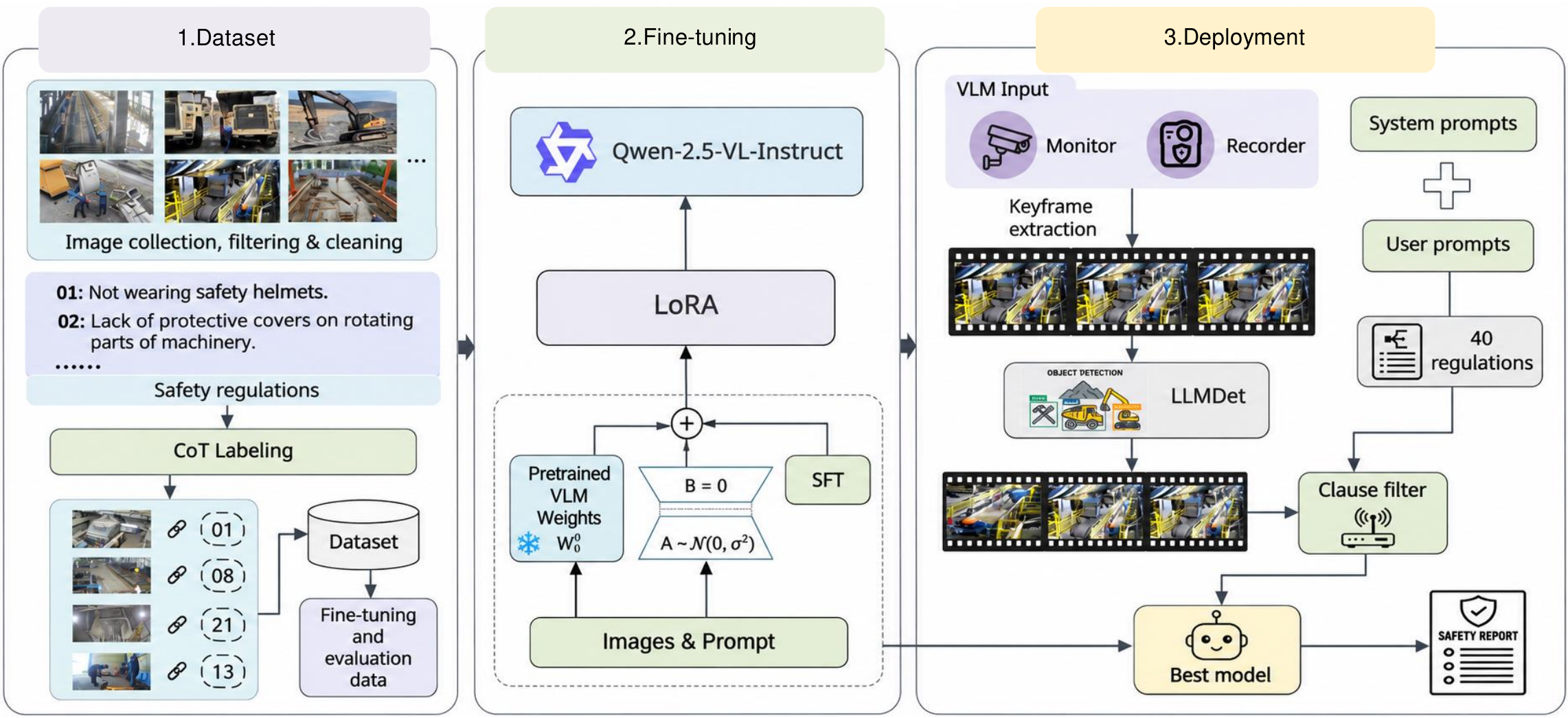}
  \caption{{\hl{Overall architecture of the proposed MonitorVLM framework. The system integrates a CF for clause filter, a BM for fine-grained worker action recognition, and a LoRA-fine-tuned VLM for comprehensive violation detection.}}}
  \label{fig:framework}
\end{figure*}

This section presents MonitorVLM, a unified framework for automated safety inspections. As illustrated in Fig.~\ref{fig:framework}, the system integrates three modules: a dynamic clause filter (CF) for selecting relevant safety regulations, a behavior magnifier (BM) for fine-grained visual enhancement, and a LoRA-fine-tuned VLM for safety analysis. By combining filtered regulations with enhanced inputs, MonitorVLM delivers precise violation detection and analytical reports. The following subsections describe the core technologies of MonitorVLM in three parts: dataset construction, model training, and model inference.

\subsection{Dataset Construction}
\label{Dataset Construction}

\begin{figure}[!htbp]
  \centering
  \includegraphics[width=0.95\columnwidth]{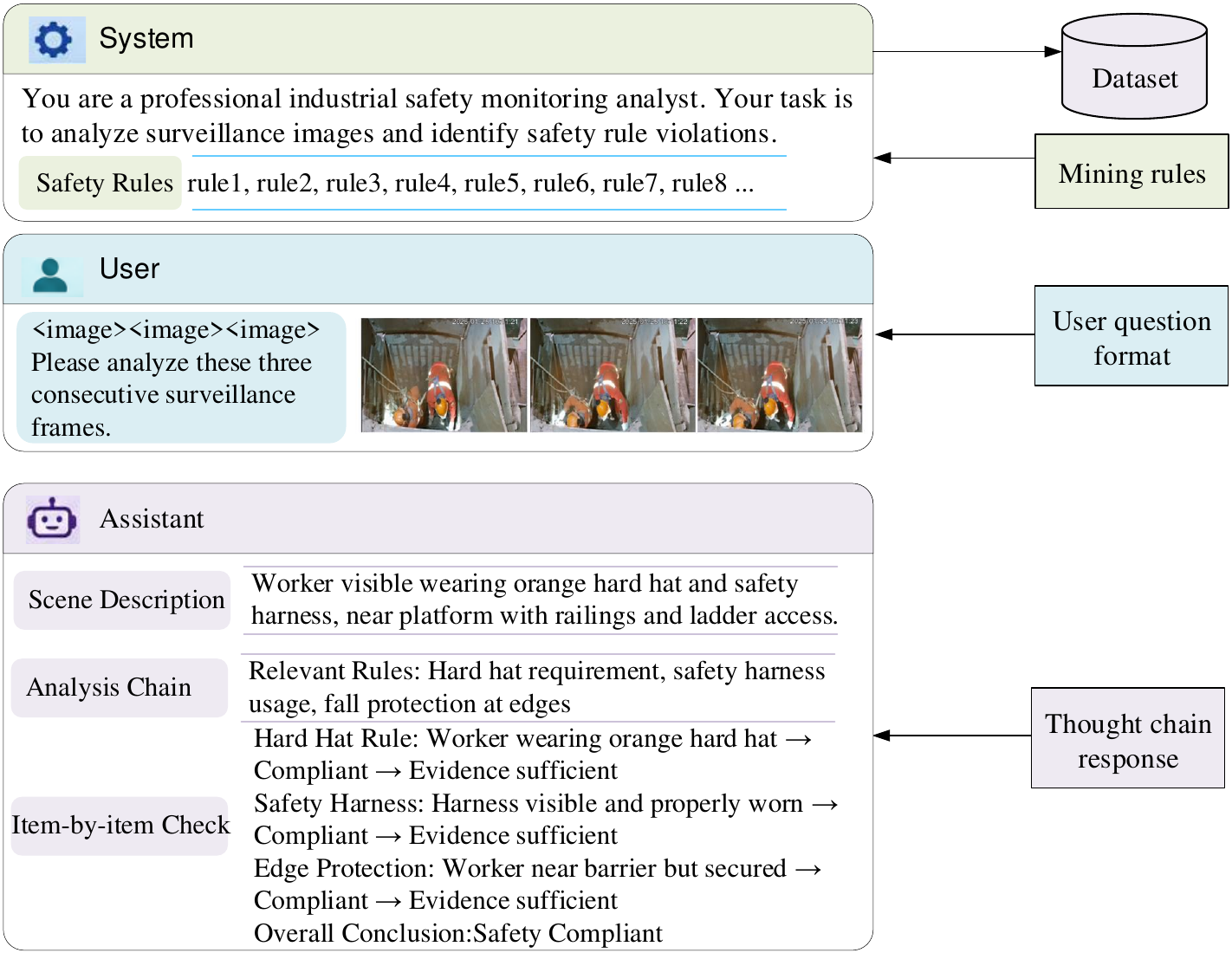}
  \caption{\hl{Structure of the constructed violation dataset. Each training instance consists of an image triplet (three temporally contiguous frames) and an instruction triplet (system prompt, user prompt, and assistant response), supporting systematic violation reasoning.}}
  \label{fig:dataset}
\end{figure}

High-quality dataset information is crucial for model fine-tuning \citep{li2021prefix}. As shown in Fig.~\ref{fig:dataset}, each training sample is organized as a paired ``image triplet'' and ``instruction triplet.'' The image triplet consists of three temporally contiguous frames captured at 1s intervals, enabling the model to perceive temporal dynamics and the evolution of violations. The instruction triplet comprises (i) a system prompt that enumerates the 40  candidate safety regulations, (ii) a user prompt requesting a detailed analytical assessment, and (iii) an assistant response that provides step-by-step reasoning followed by a systematic, clause-by-clause violation verdict.

\begin{figure}[!htbp]
  \centering
  \includegraphics[width=0.96\columnwidth]{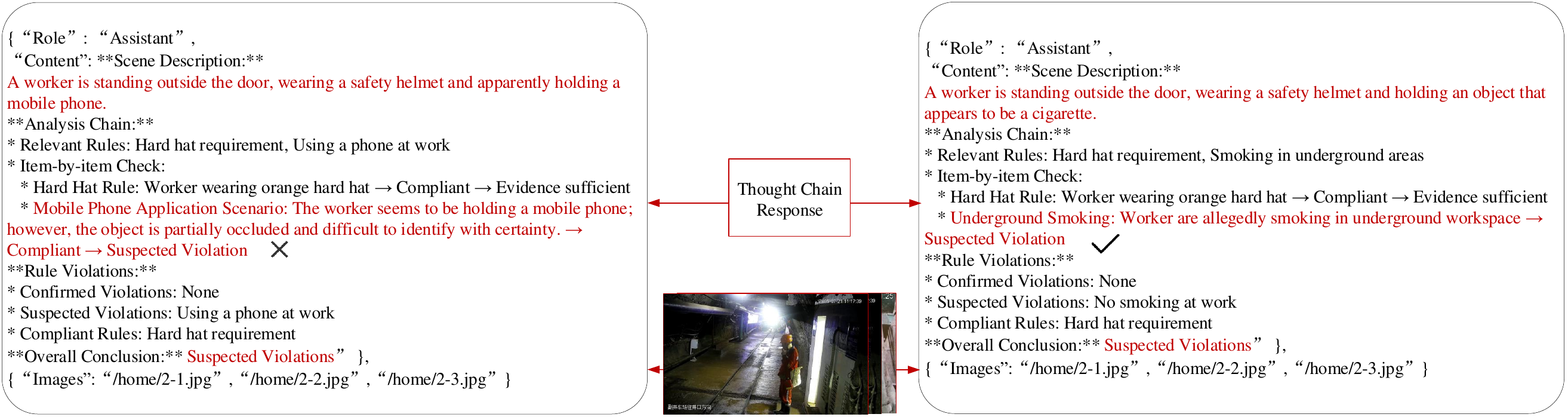}
  \caption{{\color{black}Correcting erroneous descriptions.}}
  \label{assistant}
\end{figure}

\hl{To ensure annotation quality, we adopted a two-stage expert verification procedure for the automatically generated dataset. Initially, Qwen-VL-Max generated structured ``Assistant'' responses containing scene descriptions, clause-by-clause safety assessments, and an ``Overall Conclusion''. Four certified safety experts then reviewed these annotations against the original surveillance frames and corresponding safety regulations. The review focused on two aspects: whether the final violation judgment was correct and whether the supporting descriptions accurately reflected the visual evidence and regulatory requirements. When an incorrect interpretation or conclusion was identified, one expert corrected the affected content, and another expert independently verified the revision against the original images and safety clauses. Throughout this process, the ``System'' and ``User'' prompts remained unchanged, and only erroneous content in the ``Assistant'' responses was revised to preserve the standardized annotation structure. For example, Fig.~}
\ref{assistant}
\hl{ illustrates a case in which the model incorrectly interpreted smoking as mobile phone usage, requiring corrections to both the behavioral description and violation conclusion. Overall, 30.66\% of the initial annotations required expert correction, highlighting the importance of human verification in constructing reliable domain-specific training data.}

\begin{figure}[!htbp] 
\centering 
\subfigure[]{ 
\includegraphics[width=0.96\columnwidth]{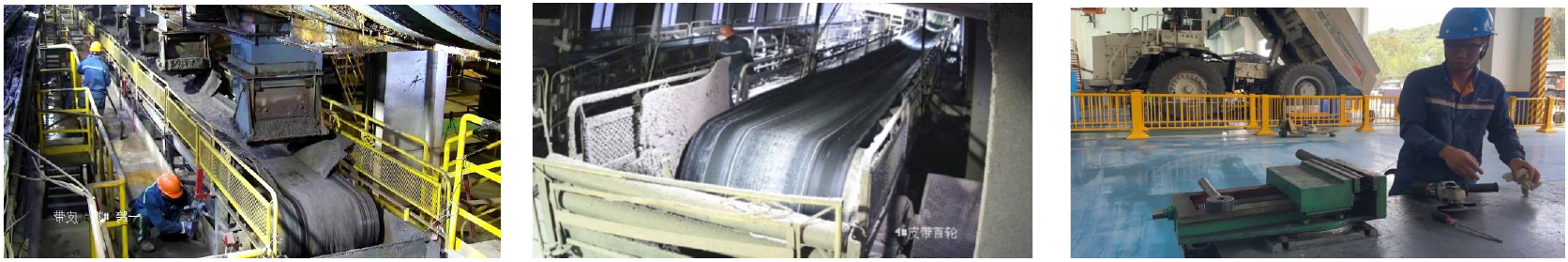}
\label{fig:Dataset1}}
\subfigure[]{
\includegraphics[width=0.96\columnwidth]{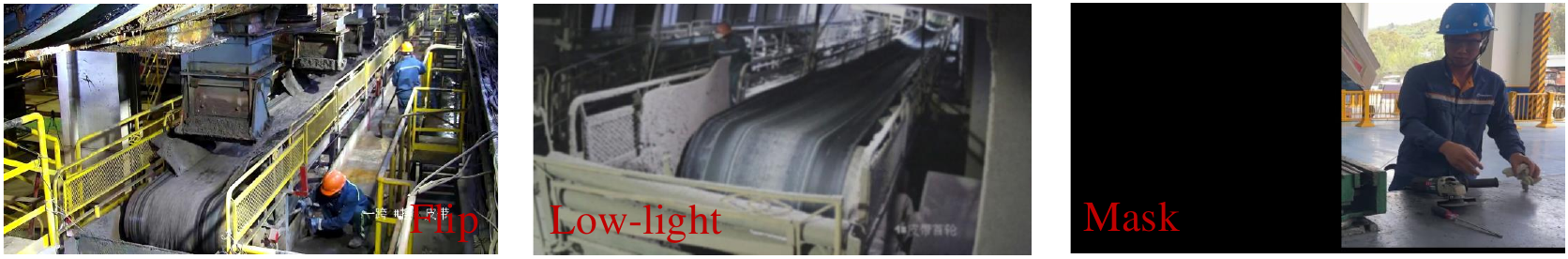}
\label{fig:Dataset2}}\\
\subfigure[]{
\includegraphics[width=0.96\columnwidth]{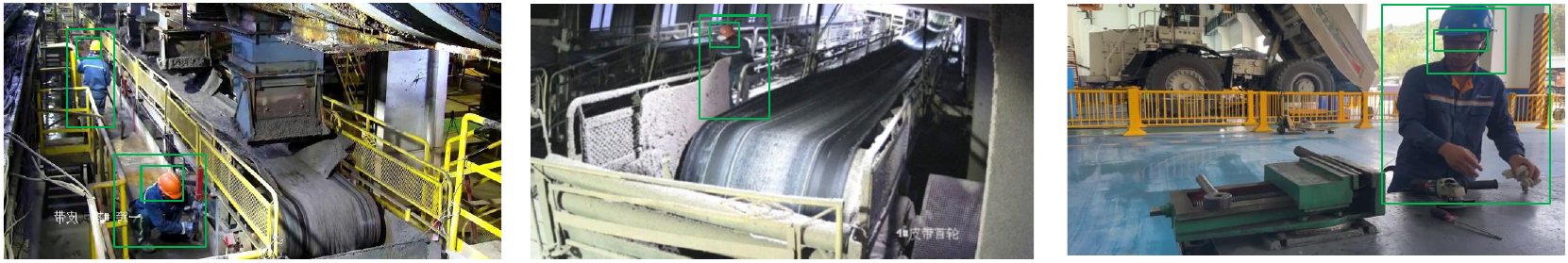}
\label{fig:Dataset3}}
\caption {Examples of dataset enrichment strategies: (a) original image, (b) image augmentation  (horizontal flipping, low-light synthesis, mask occlusion), and (c) auxiliary bounding-box annotations obtained via open-vocabulary detection.}
\end{figure}

Since the scale of violation video data in the mining industry is often limited, we introduce three image augmentation strategies to enrich the dataset and enhance model robustness: (i) horizontal flipping, which alters the spatial positions of key objects, enabling the model to better capture geometric relationships; (ii) low-light synthesis, which reduces image brightness by 20--50\% to realistically simulate the challenging illumination conditions commonly observed in mining sites or early-morning shifts; and (iii) mask occlusion, which randomly hides 10--30\% of non-critical image regions, thereby forcing the model to focus on safety-critical areas and improving detection robustness. These comprehensive transformations, illustrated in Fig.~\ref{fig:Dataset2}, significantly enrich the dataset.
In addition, we employ an open-vocabulary detector \citep{fu2025llmdet} to identify and localize common key objects in the raw images, such as worker, helmet, and safety belt. As shown in Fig.~\ref{fig:Dataset3}, the generated bounding boxes are appended as auxiliary supervision prompts, providing the model with spatial priors and further expanding the training dataset.

\subsection{Model Training}

\subsubsection{LoRA Fine-Tuning}

{\color{black}We select Qwen2.5-VL-Instruct as the backbone and adopt Low-Rank Adaptation (LoRA) for efficient domain specialization in industrial applications \citep{hu2022lora}.
    This choice is motivated by its strong vision--language performance with practical efficiency, its multi-scale variants (7B/32B/72B) for capacity analysis, and its open-weight architecture that supports domain-specific fine-tuning. In contrast to closed-source commercial models such as GPT-4o and Claude, this provides full access to model weights and architecture for industrial adaptation.} LoRA modules are strategically injected into every linear layer of both the vision encoder and the language decoder, introducing trainable low-rank matrices whereas leaving the pre-trained weights frozen.
During training on image-text pairs $(I, T)$, the standard autoregressive cross-entropy loss is formulated as:

\begin{equation}
\mathcal{L} = -\sum_{i=1}^{n} \log P(t_i | I, t_{1:i-1}; \theta_0, \theta_{\text{LoRA}})
\end{equation}
where $I$ represents the input image, $T=\{t_1, t_2, \dots, t_n\}$ denotes the target text sequence, $t_i$ is the $i$-th token, $t_{1
:i-1}$ represents the preceding token history, $\theta_0$ corresponds to the frozen pre-trained parameters, and $\theta_{\text{LoRA}}$ encompasses the trainable LoRA parameters across all linear layers.

\subsubsection{Clause filter}
\label{Router}
\begin{figure}[!hbtp]
  \centering
  \includegraphics[width=0.95\columnwidth]{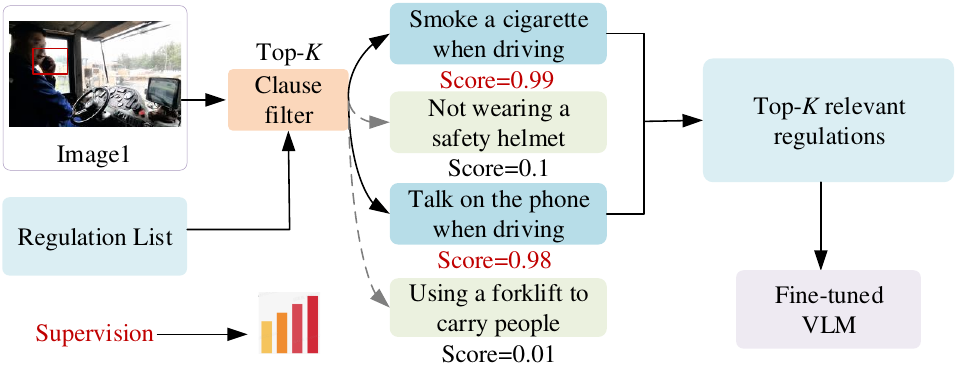}
  \caption{Illustration of the dynamic CF mechanism. The module predicts relevance scores between input frames and regulatory clauses, selecting the Top-$K$ most probable clauses for violation analysis.}
  \label{fig:router}
\end{figure}

Processing hundreds of regulatory clauses simultaneously incurs prohibitive computational cost and severe attention dilution. Analogous to the expert-selection mechanism via routers in mixture-of-experts models \citep{yang2025drivemoe}, our CF dynamically selects the most contextually relevant clauses for each mining scene, as illustrated in Fig.~\ref{fig:router}.

\begin{figure}[!htbp]
  \centering
\includegraphics[width=0.66\columnwidth]{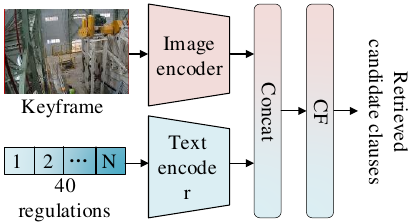}
  \caption{Dual-path network architecture of the clause filter, combining ResNet-50 visual features and BERT textual embeddings, followed by a lightweight fusion network for relevance prediction.}
  \label{fig:router_arch}
\end{figure}

During CF dataset construction, the confirmed violation labels from the expert-reviewed VLM training dataset are retained as positive labels ($1$). For the remaining image--clause pairs, Qwen-VL-Max is used to assess clause relevance and assign binary labels ($1$ for relevant and $0$ for irrelevant) to form the CF training set. As illustrated in Fig.~\ref{fig:router_arch}, the CF adopts a dual-path architecture.
A frozen ResNet-50 \citep{koonce2021resnet} encodes visual semantics, whereas a frozen BERT \citep{koroteev2021bert} encodes textual clause features.
The resulting feature vectors are concatenated and passed through a lightweight CF network for feature fusion, followed by a Sigmoid output layer that produces continuous relevance scores between 0 and 1.

The overall predicted relevance probability $p_i$ is formulated as:
\begin{equation}
    p_i = {\rm{Sigmoid}}\big( R_{\text{filter}}(I, r_i) \big)
\end{equation}
where $I$ represents the input image, $r_i$ denotes the $i$-th clause, and $R_{\text{filter}}$ is the CF network.
The CF optimization minimizes binary cross-entropy loss:
\begin{equation}
\mathcal{L}_{\text{filter}} = - \left[ y_i \cdot \log(p_i) + (1 - y_i) \cdot \log(1 - p_i) \right]
\end{equation}
where $y_i \in \{0,1\}$ is the ground-truth label. During the inference phase, we feed all regulations and key frames into CF in parallel, outputting their relevance scores in a single forward pass. This approach enables seamless expansion of the clause repository without requiring any network architecture modifications, endowing the system with high flexibility and continuous evolution capabilities.

\subsection{Model Inference}
\begin{figure}[!htbp]
  \centering
  \includegraphics[width=0.85\columnwidth]{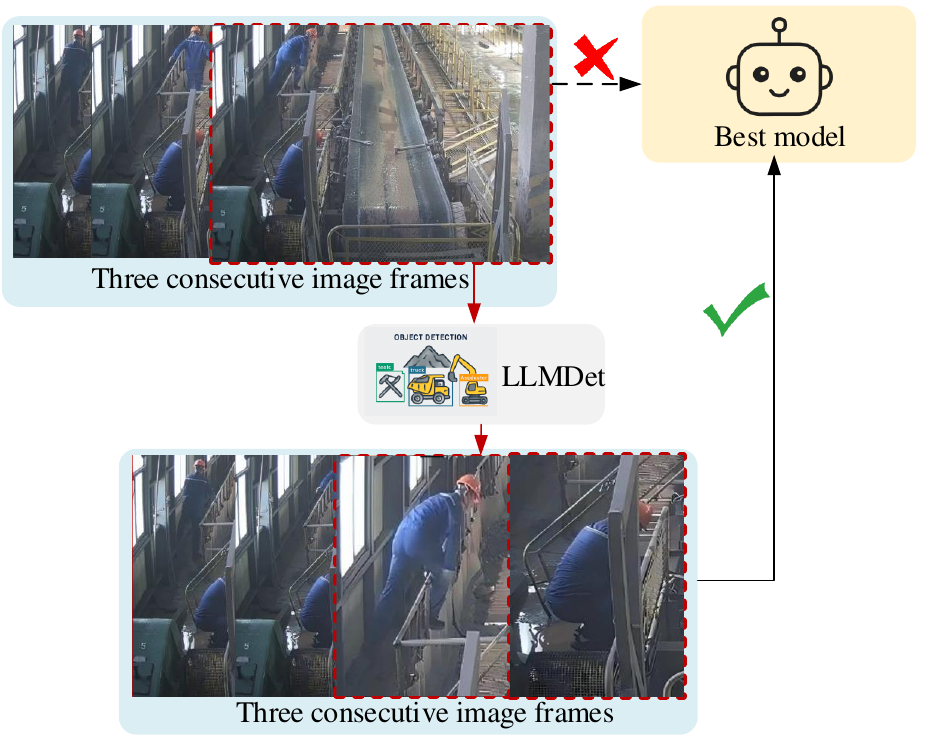}
  \caption{Processing pipeline of the Behavior Magnifier (BM). LLMDet
detects worker regions in the first frame, and the detected regions
are cropped, enlarged, and enhanced using super-resolution. The
resulting enhanced image replaces the original first frame in the
three-frame input. The other two frames remain unchanged.}
  \label{fig:object}
\end{figure}

{\color{black}
   The varying distances between cameras and workers in industrial environments can significantly degrade detection accuracy, particularly for fine-grained safety violation assessment. Small-scale objects and distant workers often suffer from insufficient visual resolution, which further exacerbates recognition errors \citep{qi2024cogcom,vukicevic2022generic}. } To address this challenge, we introduce a ``\hl{behavior magnifier}'' module for images, which applies targeted visual enhancement techniques (Fig.~\ref{fig:object}). Specifically, the LLMDet \citep{fu2025llmdet} component first performs precise personnel detection with accurate bounding box localization. Detected human regions are systematically cropped, enlarged by a factor of 2x, and enhanced using Real-ESRGAN \citep{wang2021real} super-resolution technology. BM is applied only to the first frame of each three-frame input, leaving the remaining two frames unchanged to preserve original visual evidence and mitigate the potential influence of super-resolution artifacts. The resulting enhanced image replaces the original first frame in the input sequence.

\begin{figure}[!htbp]
  \centering
  \includegraphics[width=0.76\columnwidth]{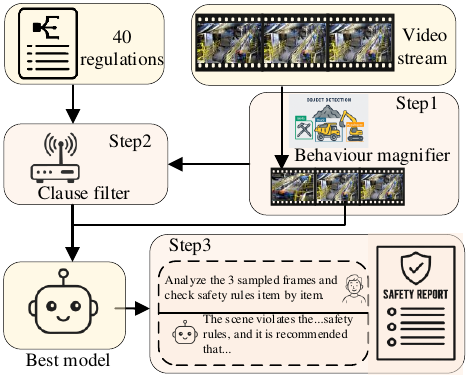}
  \caption{End-to-end inference workflow of MonitorVLM, integrating BM, CF, and the fine-tuned VLM to perform frame processing, clause filter, and clause-specific violation reporting.}
  \label{fig:Step}
\end{figure}

As shown in Fig.~\ref{fig:Step}, the offline inference pipeline consists of three stages. {\color{black}The videos are originally recorded at 30 fps. To reduce redundancy while retaining essential temporal action information, we deliberately sample the videos at 1 fps, extracting one frame per second. As a result, the three consecutive frames in our dataset reflect meaningful scene transitions and worker behavior changes, rather than near-duplicate frames.} Parallel processing is performed through a CF to identify the most relevant clauses and regions. Finally, the fine-tuned VLM integrates visual and textual inputs to assess safety violations and generate structured reports.

\section{Experiments}
\label{Experiments}
This section presents a comprehensive set of experiments that validate MonitorVLM's effectiveness in safety inspections.

\subsection{Experimental Details}
\label{Details}
Section~\ref{Dataset Construction} describes the construction of our safety violation dataset based on 40 mining safety regulations. Surveillance videos collected from 23 operational areas were processed into 3,085 training samples, each consisting of three consecutive frames and a corresponding question--answer pair. Each sample was labeled as either a violation (positive) or compliance (negative), forming Dataset I. We then apply three targeted augmentation strategies to the original data to obtain Dataset II , and incorporate open-vocabulary object-detection annotations to produce Dataset III. Finally, we merge Datasets I, II, and III to obtain a training dataset comprising 9,255 samples. \hl{The training and testing sets were partitioned at the video level rather than the frame level. Importantly, data augmentation was applied exclusively to the training set, whereas the testing set comprised 2,055 original, non-augmented samples, ensuring an independent evaluation of the proposed model.}
\hl{In Section~}\ref{Router}\hl{, the CF training dataset is constructed from 1,000 images selected exclusively from the expert-reviewed VLM training set, ensuring no overlap with the testing set. Each image is paired with all 40 safety regulations, yielding 40,000 image--clause pairs. Expert-verified violated clauses are retained as positive labels (1), and Qwen-VL-Max assigns binary relevance labels (1 for relevant and 0 for irrelevant) to the remaining image--clause pairs. This procedure preserves the expert-verified violation labels and incorporates additional relevance annotations for CF training.} For independent evaluation, 150 images are selected from the testing set and paired with all 40 safety regulations, yielding 6,000 image--clause pairs.

{\hl{We adopt the SWIFT framework~}
\citep{zhao2025swift}
\hl{ to fine-tune Qwen2.5-VL-Instruct using LoRA (}
$r=16$, $\alpha=32$
\hl{), with a learning rate of }
$1 \times 10^{-4}$
\hl{, a warm-up ratio of 0.05, and 3 training epochs on eight NVIDIA H100 GPUs. The per-device batch size is 2, with 8 gradient accumulation steps, yielding an effective batch size of 128. For the Clause Filter (CF), the ResNet-50 and BERT encoders remain frozen, and only the MLP (}
$2816 \rightarrow 1024 \rightarrow 512 \rightarrow 256 \rightarrow 1$
\hl{) is trained using Adam (}
$1 \times 10^{-3}$
\hl{), with ReLU hidden activations and a Sigmoid output. We compare MonitorVLM against eight baselines: Qwen2.5-VL-Instruct (7B/32B/72B)~}
\citep{bai2025qwen2}
\hl{, LLaVA-OneVision (7B/72B)~}
\citep{li2024llava}
\hl{, GPT-4o~}
\citep{yan2025gpt}
\hl{, Gemini-2.5~}
\citep{comanici2025gemini}
\hl{, and Claude-3.7-Sonnet~}
\citep{anderson2025comparative}
\hl{. All open-source baselines are fine-tuned on eight NVIDIA H100 GPUs under the same training configuration, whereas proprietary models are evaluated directly through APIs. All models follow the same prompting format in Fig.~}
\ref{fig:dataset}
\hl{, and prediction correctness is assessed against ground-truth labels.}}

 \hl{We adopt a clause-constrained sample-level evaluation protocol. Each violating sample has one annotated violated clause, and a prediction is considered correct only when the corresponding clause is accurately identified. Both confirmed and suspected violations are treated as violation predictions. The metrics are defined as follows:}
\begin{equation}
\mathrm{Precision} =
\frac{\mathrm{TP}}{\mathrm{TP}+\mathrm{FP}}
\label{eq:binary_precision}
\end{equation}

\begin{equation}
\mathrm{Recall} =
\frac{\mathrm{TP}}{\mathrm{TP}+\mathrm{FN}}
\label{eq:binary_recall}
\end{equation}

\begin{equation}
\mathrm{F1} =
2 \cdot
\frac{\mathrm{Precision}\cdot\mathrm{Recall}}
{\mathrm{Precision}+\mathrm{Recall}}
\label{eq:binary_f1}
\end{equation}

\begin{equation}
\mathrm{Accuracy} =
\frac{\mathrm{TP}+\mathrm{TN}}
{\mathrm{TP}+\mathrm{TN}+\mathrm{FP}+\mathrm{FN}}
\label{eq:binary_accuracy}
\end{equation}
\hl{TP denotes correctly identified violations with matching clauses, FN denotes missed violations or incorrect clause matches, FP denotes compliant samples incorrectly flagged as violating, and TN denotes correctly identified compliant samples. Incorrect clause matches are counted only as FN. Thus, the reported metrics follow a task-specific counting scheme rather than conventional binary classification criteria.}

\subsection{Overall Performance}
\label{Overall_Performance}
\begin{figure}[!htbp]
\centering
\subfigure[Training set]{
    \begin{tabular}{cc}
    \includegraphics[width=0.40\columnwidth]{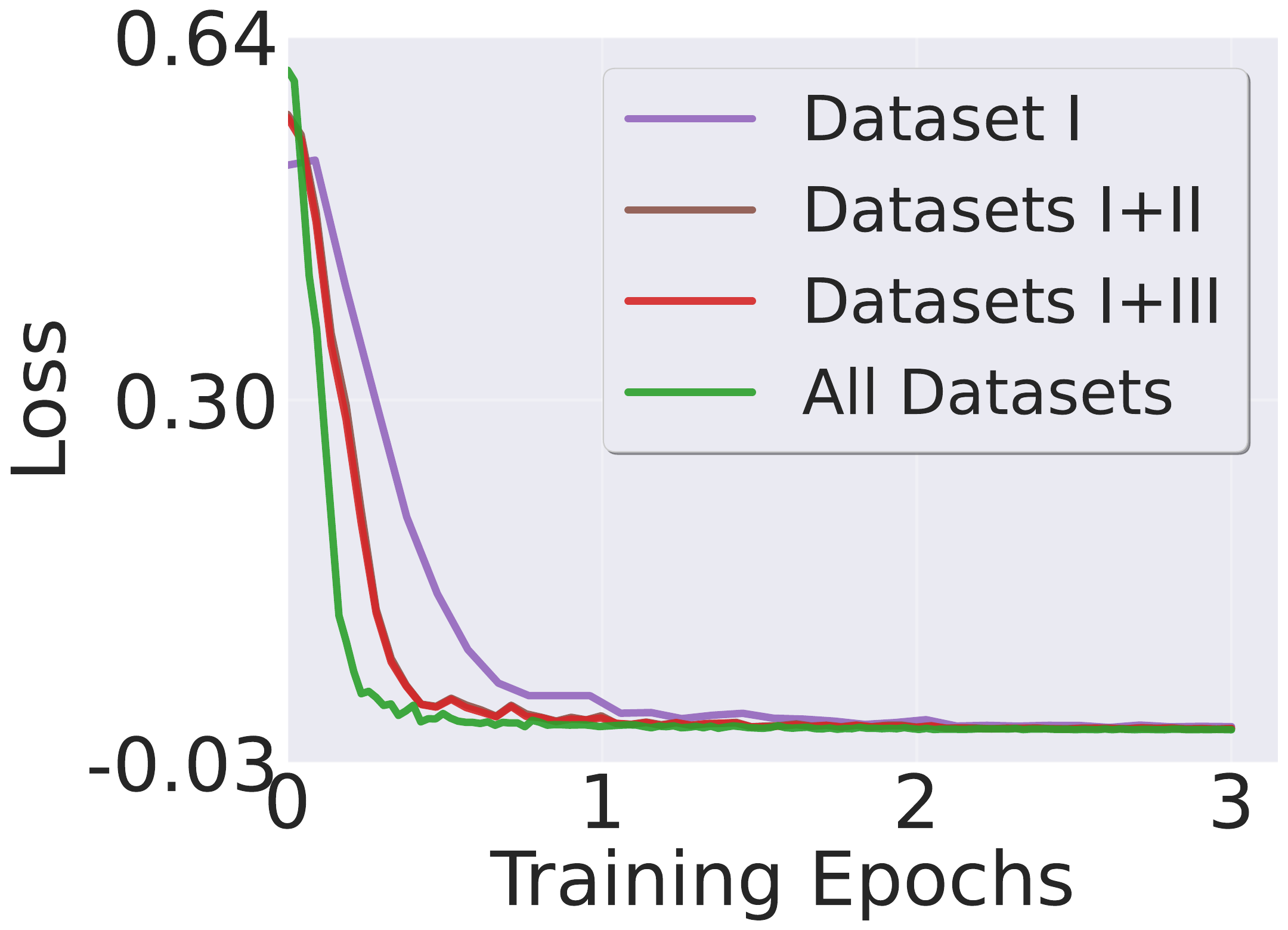} &
    \includegraphics[width=0.40\columnwidth]{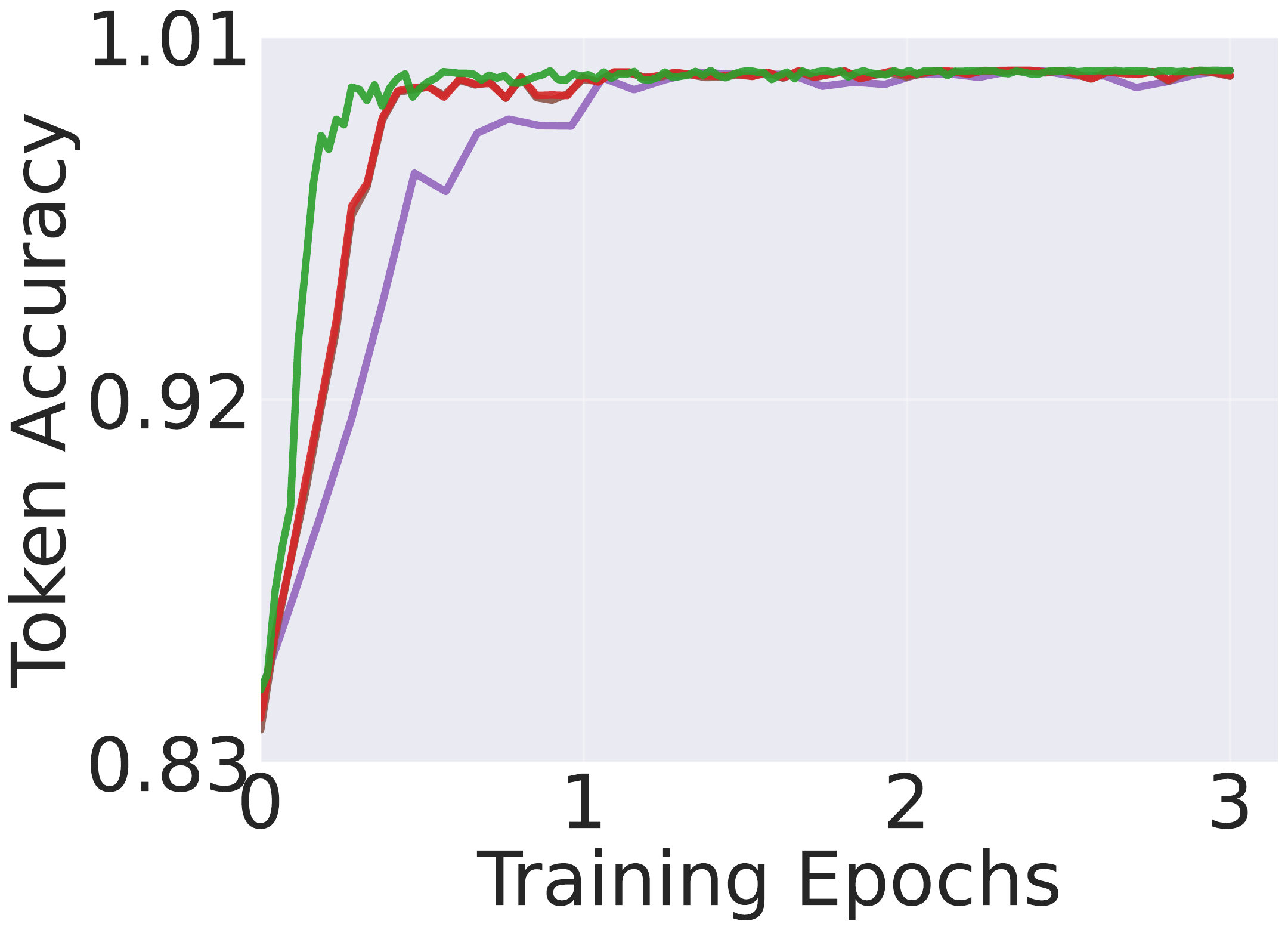}
    \end{tabular}
    \label{fig:dataset1}}\\
\subfigure[\hl{Validation set}]{
    \begin{tabular}{cc}
    \includegraphics[width=0.40\columnwidth]{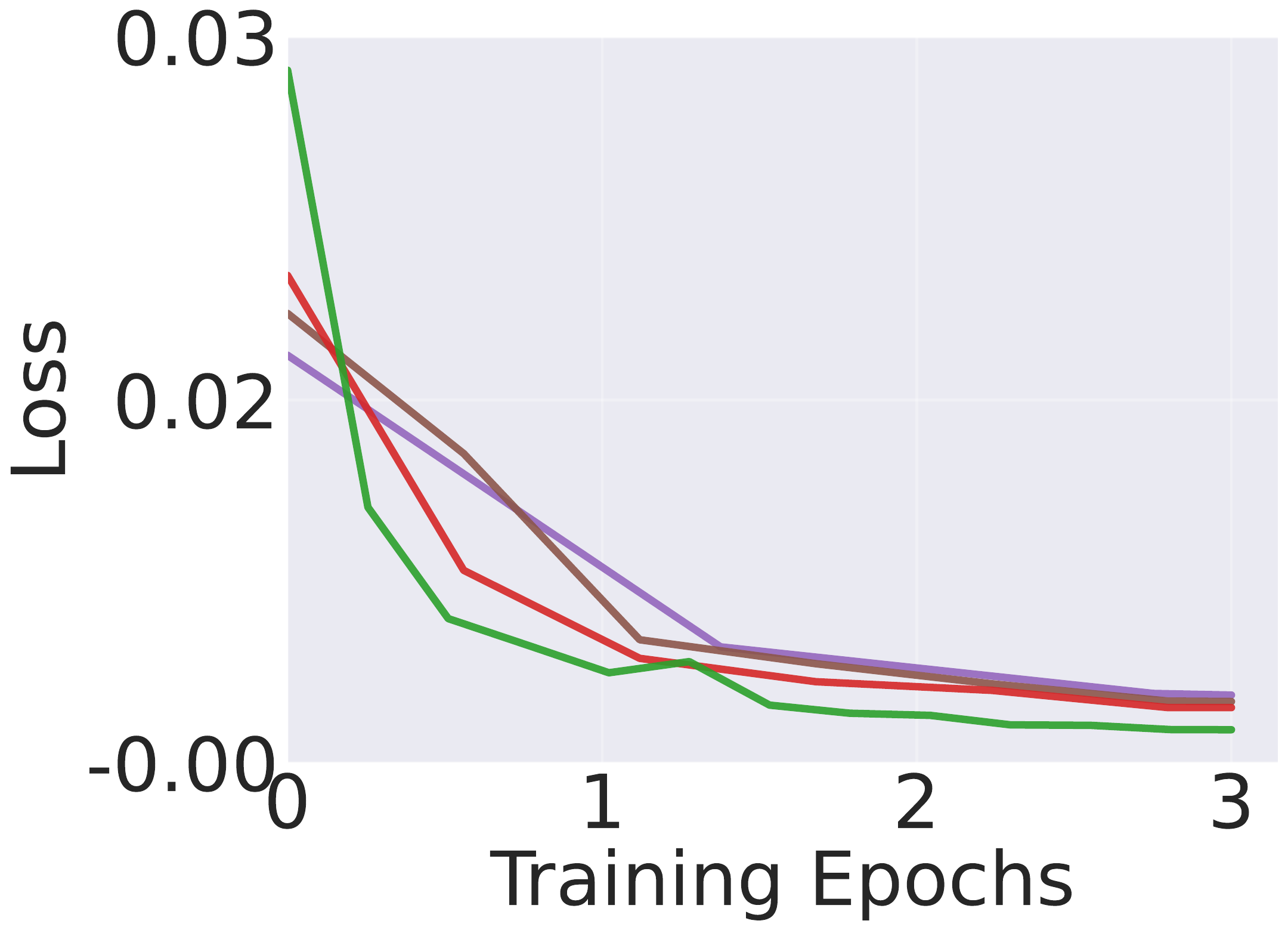} &
    \includegraphics[width=0.40\columnwidth]{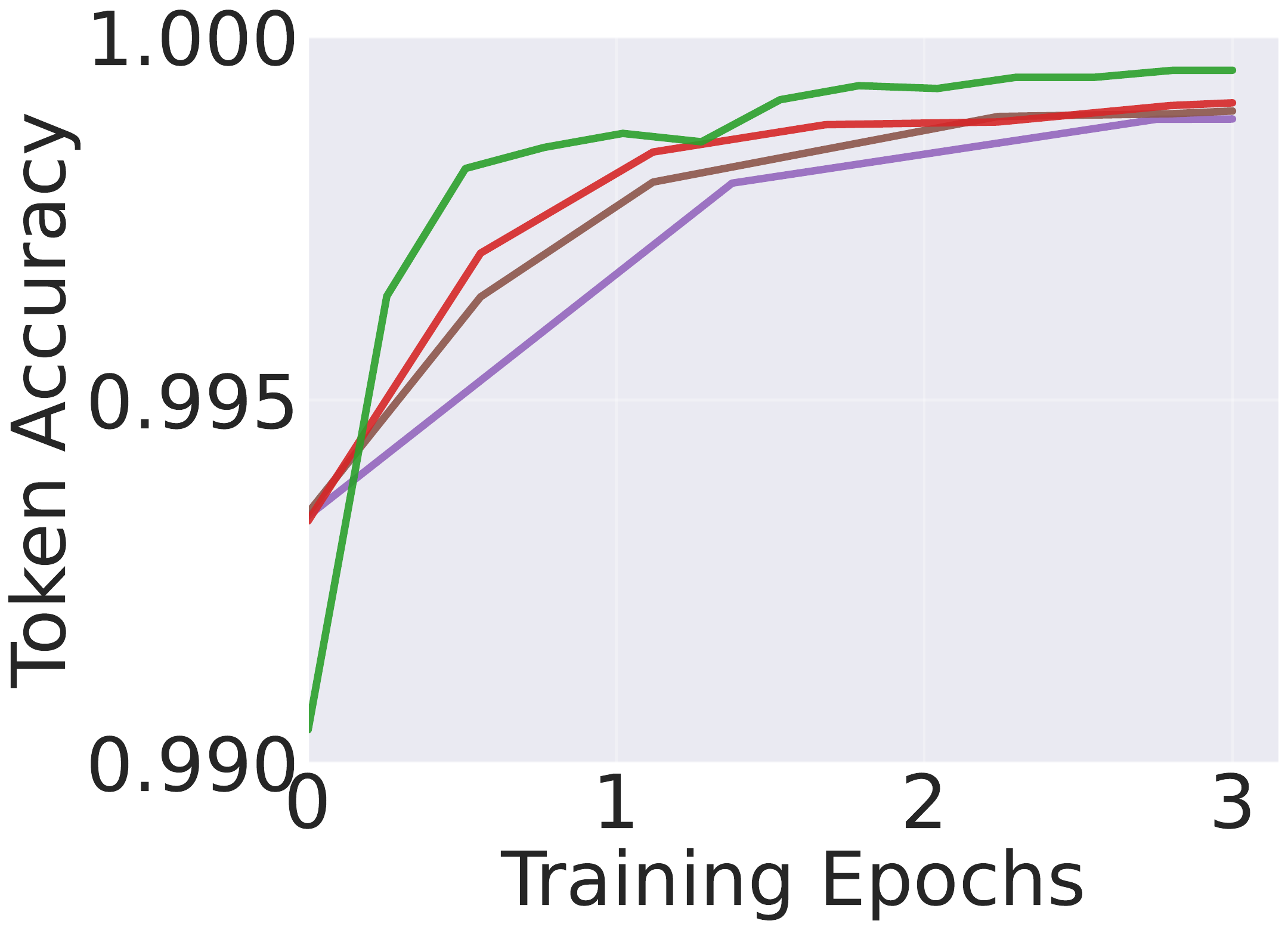}
    \end{tabular}
    \label{fig:dataset2}}
\caption{Training and evaluation curves of Qwen2.5-VL-72B fine-tuned on different dataset configurations. Token accuracy denotes the proportion of correctly predicted tokens during validation.}
\label{fig.results}
\end{figure}

To assess the impact of different datasets on fine-tuning, we conduct domain-specific adaptation of Qwen2.5-VL-72B using four configurations: Dataset~I, Datasets~I+II, Datasets~I+III, and the full combination of all three datasets. \hl{We reserve 1\% of the training data as a validation set to  monitor fine-tuning. }The training curves in Fig.~\ref{fig.results} show stable convergence across all settings. On the validation set, the model fine-tuned on all datasets achieves the highest token prediction accuracy, and we refer to this variant as MonitorVLM-72B-basic. In addition, MonitorVLM-72B denotes the complete system, which extends MonitorVLM-72B-basic by incorporating the CF (responsible for selecting the Top-5 most relevant clauses for the VLM) and the BM ( designed to enhance the recognition of fine-grained worker actions).

Table~\ref{tab:model_performance} reports the detailed performance of all baseline models on the testing set. The results indicate that fine-tuning substantially improves the effectiveness of both the Qwen2.5-VL-basic variants (7B, 32B, and 72B) {\color{black}and LLaVA-OneVision variants (7B, 72B)} compared with their unfine-tuned counterparts. For instance, relative to Qwen2.5-VL-72B, the fine-tuned MonitorVLM-72B-basic achieves absolute gains of 16.16, 20.89, and 18.91 percentage points in precision, recall, and F1 score, respectively. {\color{black}Similarly, LLaVA-OneVision-72B-basic shows marked improvement over its base model, with gains of 19.15, 15.89, and 17.45 percentage points in precision, recall, and F1 score, respectively, further demonstrating the effectiveness of domain-specific fine-tuning. Under the same training setting, MonitorVLM-72B-basic outperforms LLaVA-OneVision-72B-basic by 5.84, 20.12, and 14.44 percentage points in precision, recall, and F1 score, respectively, validating the superiority of our proposed approach.} Furthermore, when the CF and BM modules are integrated, MonitorVLM-72B achieves additional improvements of 3.57 and 7.59 percentage points in precision and recall, respectively, underscoring the effectiveness of the proposed modular enhancements.

\begin{table*}[htbp]
\centering
\caption{\hl{Model Performance Comparison.}}
\label{tab:model_performance}
\setlength{\tabcolsep}{7pt}
\begin{tabular}{lccc}
\toprule
\hl{\textbf{Model}} & \hl{Precision} & \hl{Recall} & \hl{F1 score} \\
\midrule
\hl{Claude-3.7-Sonnet} & \hl{75.43\%} & \hl{55.49\%} & \hl{63.94\%} \\
\hl{GPT-4o} & \hl{77.91\%} & \hl{61.25\%} & \hl{68.58\%} \\
\hl{Gemini-2.5} & \hl{76.39\%} & \hl{59.14\%} & \hl{66.67\%} \\
\hl{LLaVA-OneVision-7B} & \hl{45.19\%} & \hl{23.14\%} & \hl{30.60\%} \\
\hl{LLaVA-OneVision-72B} & \hl{64.50\%} & \hl{45.99\%} & \hl{53.69\%} \\
\hl{Qwen2.5-VL-7B} & \hl{62.53\%} & \hl{31.93\%} & \hl{42.27\%} \\
\hl{Qwen2.5-VL-32B} & \hl{69.74\%} & \hl{53.16\%} & \hl{60.34\%} \\
\hl{Qwen2.5-VL-72B} & \hl{73.33\%} & \hl{61.11\%} & \hl{66.67\%} \\

\midrule

\hl{Qwen2.5-VL-72B (Dataset I)} & \hl{78.96\%} & \hl{65.19\%} & \hl{71.42\%} \\
\hl{Qwen2.5-VL-72B (Datasets I+II)} & \hl{89.13\%} & \hl{79.61\%} & \hl{84.10\%} \\
\hl{Qwen2.5-VL-72B (Datasets I+III)} & \hl{84.17\%} & \hl{68.07\%} & \hl{75.27\%} \\

\hl{LLaVA-OneVision-7B-basic} & \hl{70.52\%} & \hl{46.77\%} & \hl{56.24\%} \\
\hl{LLaVA-OneVision-72B-basic} & \hl{83.65\%} & \hl{61.88\%} & \hl{71.14\%} \\
\hl{MonitorVLM-7B-basic} & \hl{80.28\%} & \hl{68.42\%} & \hl{73.88\%} \\
\hl{MonitorVLM-32B-basic} & \hl{82.37\%} & \hl{72.93\%} & \hl{77.36\%} \\
\hl{MonitorVLM-72B-basic} & \hl{89.49\%} & \hl{82.00\%} & \hl{85.58\%} \\

\midrule

\hl{MonitorVLM-72B-basic w/ CF} & \hl{89.95\%} & \hl{82.49\%} & \hl{86.06\%} \\
\hl{MonitorVLM-72B-basic w/ BM} & \hl{93.04\%} & \hl{89.31\%} & \hl{91.14\%} \\
\hl{MonitorVLM-72B} & \hl{93.06\%} & \hl{89.59\%} & \hl{91.29\%} \\

\bottomrule
\end{tabular}
\vspace{5pt}
\noindent
\footnotesize
\end{table*}

\subsection{Ablation study}
We next analyze the impact of data augmentation, CF, and BM on the performance of MonitorVLM-72B.
\subsubsection{Impact of data augmentation}
\hl{As shown in Table~}
\ref{tab:model_performance}
\hl{, we compare Qwen2.5-VL-72B fine-tuned on four dataset configurations: Dataset I, Datasets I+II, Datasets I+III, and the combination of all three datasets. Compared with Dataset I, image augmentation (Datasets I+II) improves precision, recall, and F1 score by 10.17, 14.42, and 12.68 percentage points, respectively, demonstrating its substantial contribution to safety violation recognition through increased visual diversity. Auxiliary bounding-box annotations (Datasets I+III) also improve these metrics by 5.21, 2.88, and 3.85 percentage points, respectively, by providing complementary spatial information. These results indicate that image augmentation contributes more significantly to performance improvement, and both strategies offer complementary benefits. Combining both strategies achieves the best performance, with precision, recall, and F1 score reaching 89.49\%, 82.00\%, and 85.58\%, respectively.}

\subsubsection{Effectiveness of the clause filter}
\label{clause filter}
\begin{figure}[!htbp]
\centering
\subfigure[Coverage, recall, and precision at different Top-$K$ values.]{
    \includegraphics[width=0.76\columnwidth]{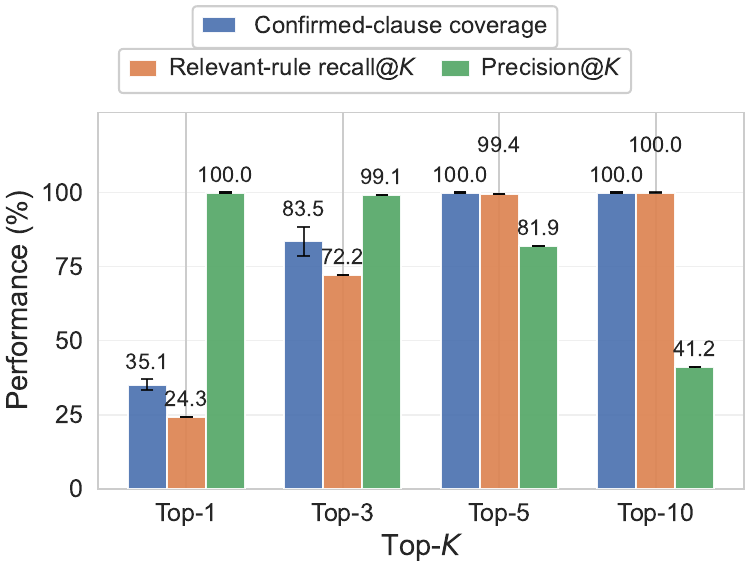}
    \label{fig.topk_com}
}
\subfigure[Inference time and Top-5 confirmed-clause coverage under regulation-repository expansion.]{
    \includegraphics[width=0.78\columnwidth]{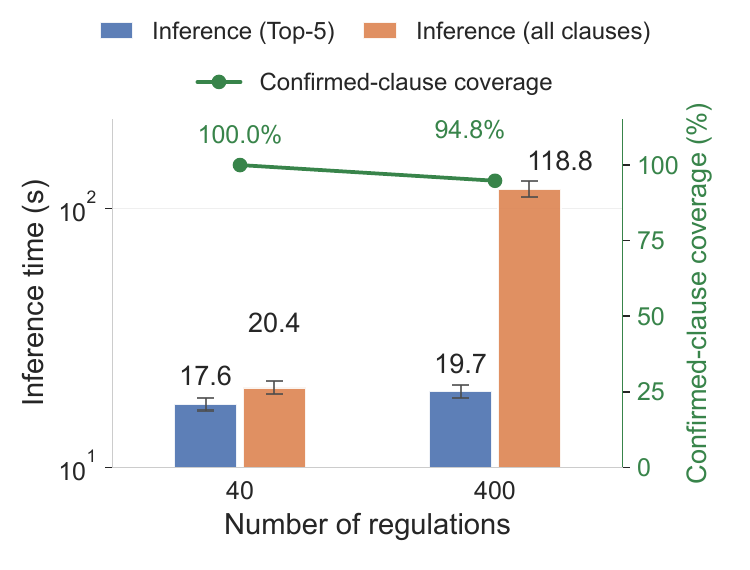}
    \label{fig.time_com}}
\caption{\hl{Evaluation of the Clause Filter (CF). All experiments are conducted on eight NVIDIA H100 GPUs.}}
\end{figure}

\hl{Fig.~}
\ref{fig.topk_com}
\hl{ evaluates three complementary metrics of the Clause Filter (CF) under different Top-}
$K$
\hl{ settings: ground-truth violation coverage, relevant-rule Recall@}
$K$
\hl{, and Precision@}
$K$
\hl{. Precision reaches 100\% at }
$K=1$
\hl{, indicating that the top-ranked clause is relevant for every evaluated sample. As }
$K$
\hl{ increases, precision decreases from 99.1\% at }
$K=3$
\hl{ to 81.9\% at }
$K=5$
\hl{ and 41.2\% at }
$K=10$
\hl{. This decline is mainly attributed to the limited number of relevant regulations associated with each scene. When }
$K$
\hl{ exceeds the number of relevant clauses, additional irrelevant regulations are included, leading to a decline in precision. Nevertheless, maintaining high coverage and recall is essential for CF because omitted candidate safety regulations cannot be recovered by the downstream VLM. Ground-truth violation coverage increases from 35.1\% at }
$K=1$
\hl{ to 83.5\% at }
$K=3$
\hl{, reaching 100\% at }
$K=5$
\hl{. Similarly, Recall@}
$K$
\hl{ increases from 24.3\% to 72.2\% and 99.4\%, respectively. Increasing }
$K$
\hl{ to 10 improves recall by only 0.6 percentage points but nearly halves precision and doubles the number of clauses passed to the VLM. Therefore, we adopt }
$K=5$
\hl{ as the smallest setting achieving complete ground-truth violation coverage within the evaluated 40-clause set, providing a favorable balance between retrieval effectiveness and computational efficiency. Furthermore, the CF achieves a micro ROC-AUC of 0.9993 over the complete image--clause score matrix, demonstrating strong discrimination between relevant and irrelevant clauses.}

\begin{figure*}[!htpb]        
  \centering
  \includegraphics[width=0.95\textwidth]{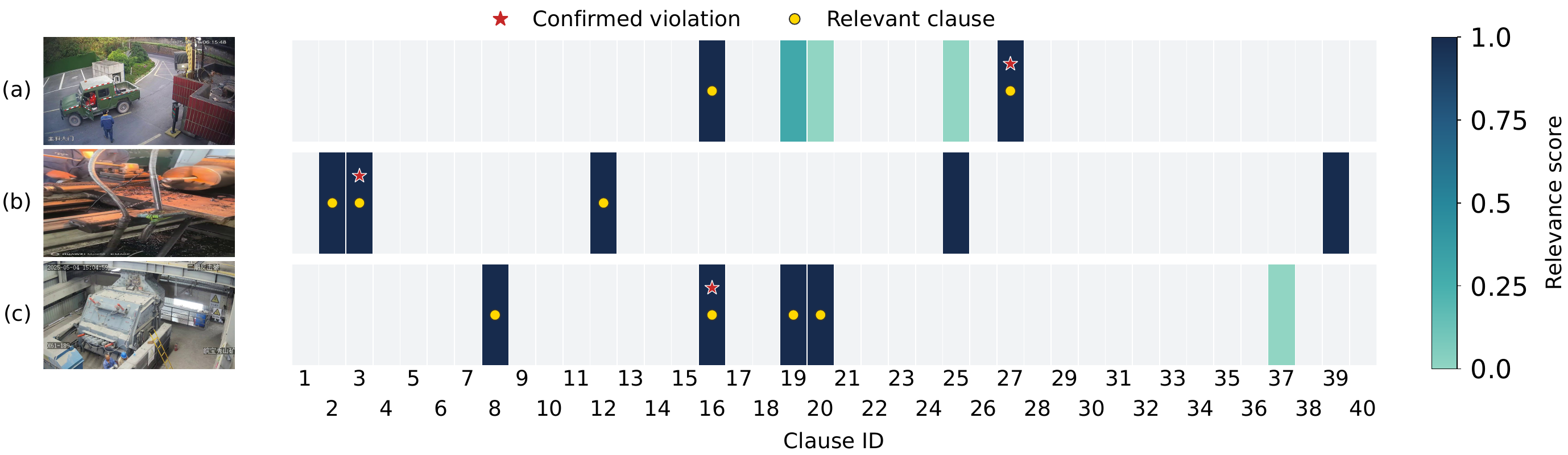}
\caption{Qualitative examples of multi-label clause retrieval using CF. Darker colors indicate higher predicted relevance scores. Yellow dots mark reference relevant clauses, and red stars indicate reference confirmed violations. Light gray cells denote scores omitted for clarity.}

\label{fig.cf_examples}

\end{figure*}

\hl{Fig.~}\ref{fig.cf_examples}
\hl{presents three illustrative examples of multi-label clause retrieval from the 40-clause repository. CF assigns a relevance score to each regulation, with darker colors indicating higher scores. Yellow dots mark reference relevant clauses, while red stars indicate reference confirmed violations. These examples illustrate that multiple regulations can be relevant to the same scene, but a relevant regulation is not necessarily violated. CF retrieves potentially applicable regulations, and the downstream VLM subsequently assesses whether the visual evidence supports a violation.}

\hl{As shown in Tables~}
\ref{tab:model_performance} and \ref{tab:different_top_k}
\hl{, incorporating CF into the fine-tuned baseline improves precision, recall, and F1-score by 0.46, 0.49, and 0.48 percentage points, respectively. Table~}
\ref{tab:different_top_k}
\hl{ compares the detection performance of MonitorVLM-72B under different Top-}
$K$
\hl{ settings. At }
$K=3$
\hl{, performance decreases substantially because relevant safety clauses may be excluded, whereas }
$K=5$
\hl{ achieves high precision and recall with complete ground-truth violation coverage. As shown in Fig~}
\ref{fig.time_com}
\hl{, incorporating CF maintains comparable detection performance and reduces inference time from 20.4~s to 17.6~s, representing a 13.73\% reduction.}
\hl{Furthermore, Fig.~}
\ref{fig.time_com}
\hl{ evaluates the inference time and confirmed-clause coverage as the regulation repository expands from 40 to 400 clauses, with }
$K$
\hl{ fixed at 5. With CF, the average inference time increases only slightly from 17.6~s to 19.7~s, compared with 118.8~s when directly processing all 400 clauses. Meanwhile, confirmed-clause coverage decreases from 100.0\% to 94.8\%. This reduction may be attributed to semantic similarities among certain regulations, as newly introduced clauses can receive higher relevance scores and push the ground-truth clause outside the Top-5 selections. Although increasing }
$K$
\hl{ can improve coverage within larger regulatory databases, it also introduces additional inference overhead.
For our current task involving 40 safety regulations, }
$K=5$
\hl{ achieves 100\% confirmed-clause coverage with low inference latency, providing an effective balance between retrieval coverage and computational efficiency.}

\begin{table*}[!htbp]
\centering
\caption{\hl{Performance of MonitorVLM-72B under varying $K$.}}
\label{tab:different_top_k}
\begin{tabular}{ccccc}
\toprule
Parameter of CF & Precision & Recall & F1 score & Inference time (s) \\
\midrule
$K=1$  & 35.93\% & 24.96\% & 29.46\% & 17.46 \\
$K=3$  & 75.60\% & 70.60\% & 73.02\% & 17.51 \\
$K=5$  & \text{93.06\%} & \text{89.59\%} & \text{91.29\%} & 17.59 \\
$K=10$ & 93.13\% & 89.59\% & 91.33\% & 18.03 \\
\bottomrule
\end{tabular}
\end{table*}

\subsubsection{Behavior magnifier effect demonstration}
\begin{figure}[!htpb]
\centering
\subfigure[Performance comparison among original inputs, cropped inputs, Lanczos 2\ensuremath{\times} upscaling, and Real-ESRGAN 2\ensuremath{\times} super-resolution.]{
    \includegraphics[width=0.89\columnwidth]{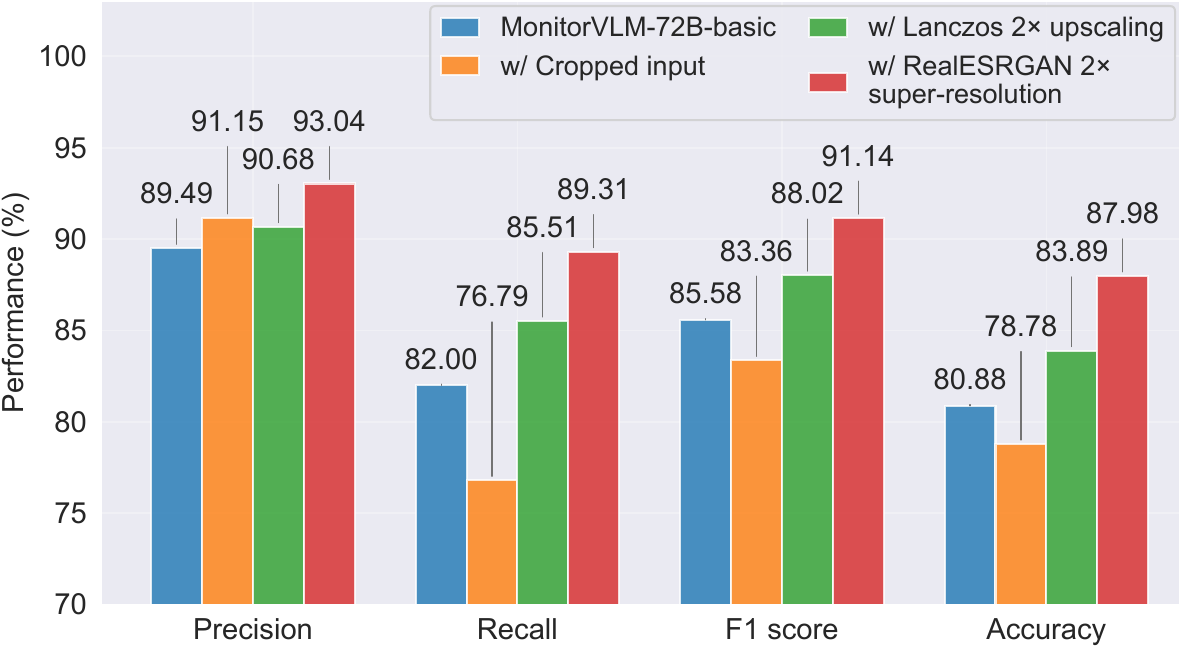}
    \label{fig.success}}
\subfigure[Representative successful and unsuccessful cases of safety helmet violation detection with BM.]{
    \includegraphics[width=0.72\columnwidth]{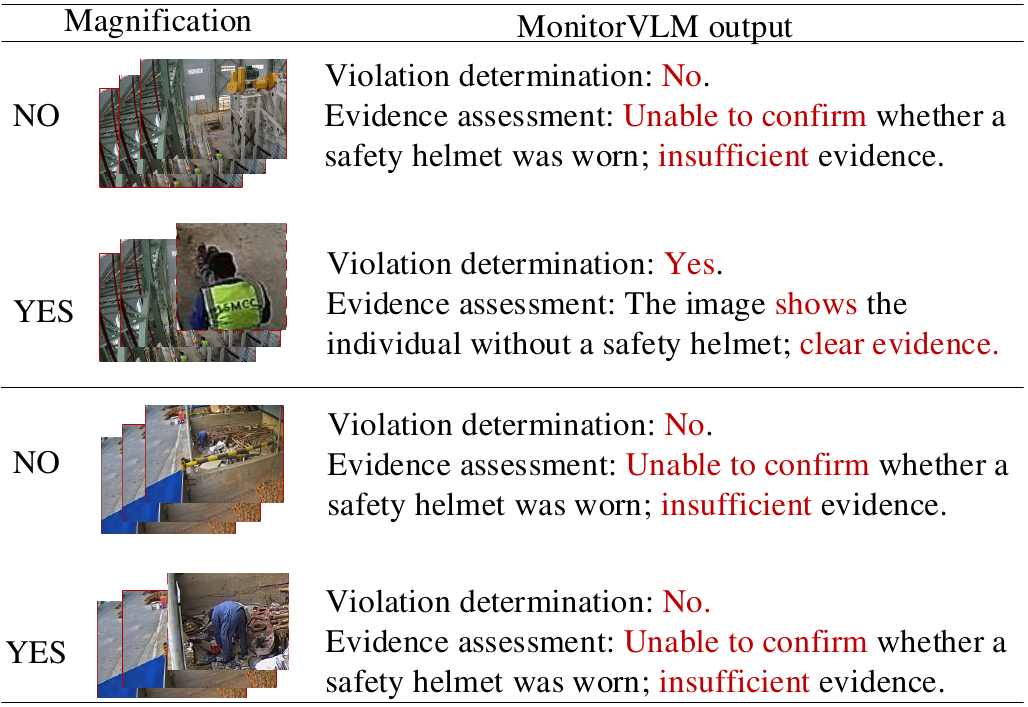}
    \label{fig.failure}}
\caption{{\color{black}Case studies demonstrating the effect of the BM in enhancing violation detection.}}
\end{figure}

\hl{The Behavior Magnifier (BM) aims to improve the recognition of distant workers and small safety-related objects by enhancing only the first frame of each three-frame input, leaving the other two frames unchanged. To evaluate its effectiveness, we compare four configurations: the original input, cropped input, Lanczos 2\ensuremath{\times} upscaling, and Real-ESRGAN 2\ensuremath{\times} super-resolution. As shown in Fig.~}
\ref{fig.success}
\hl{, Real-ESRGAN achieves the best performance, improving precision, recall, F1 score, and accuracy by 3.55, 7.31, 5.56, and 7.10 percentage points, respectively, compared with MonitorVLM-72B-basic. Cropping alone, which preserves the original pixels without upscaling, does not improve accuracy, suggesting that isolating person regions is insufficient when visual details remain limited. Lanczos 2\ensuremath{\times} upscaling improves performance through interpolation, whereas Real-ESRGAN 2\ensuremath{\times} super-resolution yields larger gains, potentially by enhancing visual cues associated with distant workers and small safety-related objects.}
\hl{Fig.~}
\ref{fig.failure}
\hl{ presents two representative helmet detection cases. In the first case, BM enables correct violation identification by enhancing previously insufficient visual details. In the second case, the model remains unable to determine helmet compliance due to an obstructed camera viewpoint. Furthermore, super-resolution may reconstruct misleading details of small safety-related objects, such as helmets, phones, or safety harnesses, potentially causing incorrect predictions. Retaining two unenhanced frames provides additional original visual evidence but does not eliminate this risk. Future work will explore alternative enhancement and context-preserving strategies to improve BM reliability.}

\subsection{Temporal Analysis and Deployment Feasibility}

\subsubsection{Impact of Temporal Information on Violation Detection}
\begin{figure}[!htbp]
  \centering
  \includegraphics[width=0.96\columnwidth]{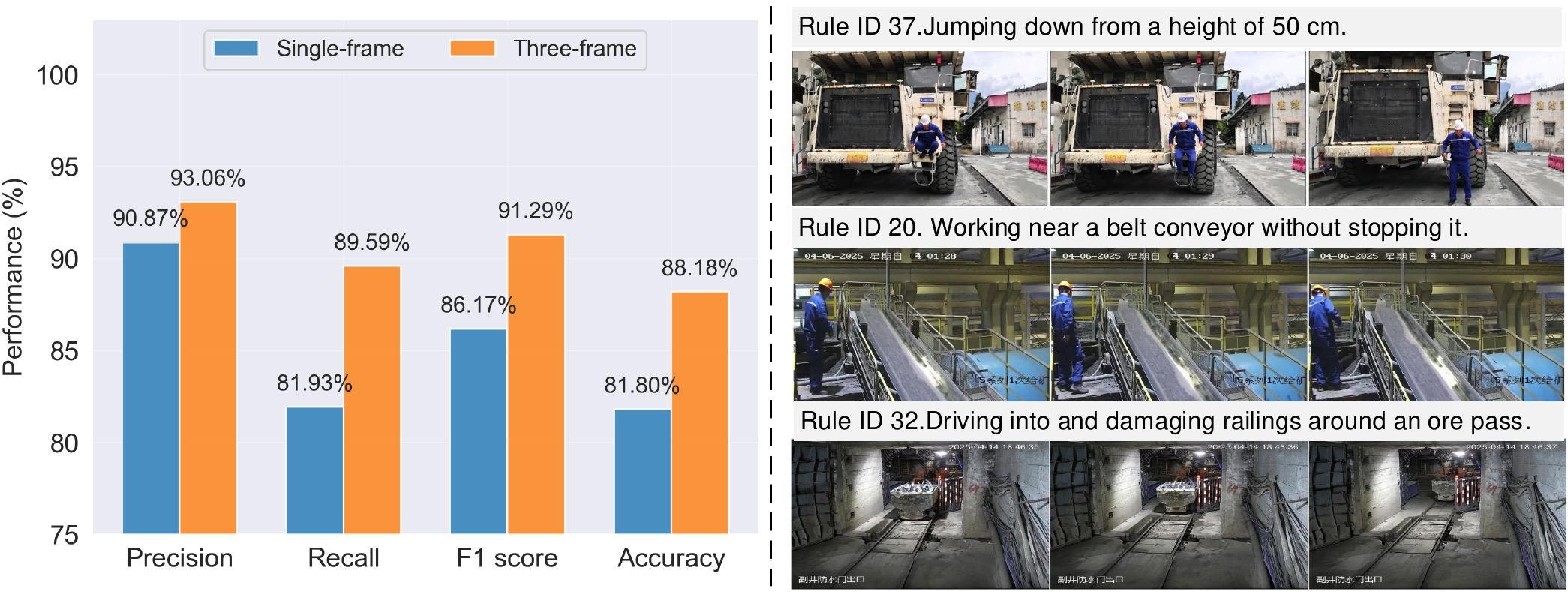}
  \caption{\hl{Comparison of single-frame and three-frame inputs for safety violation detection: (a) precision, recall, F1 score, and accuracy; (b) representative cases illustrating the importance of temporal information for recognizing dynamic safety violations.}}
  \label{fig.temporal_comparison}
\end{figure}

\hl{To evaluate the contribution of temporal information to safety violation detection, we compare single-frame inputs with three-frame inputs sampled at one-second intervals under identical experimental conditions. As shown in Fig.~}
\ref{fig.temporal_comparison}
\hl{, three-frame inputs improve precision, recall, F1 score, and accuracy by 2.19, 7.66, 5.12, and 6.38 percentage points, respectively. The substantial improvement in recall indicates that temporal information effectively reduces missed violations. We further present three representative cases involving jumping from a height (Rule 37), working near an operating belt conveyor (Rule 20), and damaging protective railings with a vehicle (Rule 32). In these scenarios, a single frame cannot reliably capture the progression of worker actions or changes in equipment status, making it difficult to determine whether a violation has occurred. In contrast, three consecutive frames provide temporal evidence of these changes, enabling more accurate recognition of dynamic safety violations.}

{\color{black}
\subsubsection{Inference Performance on Different GPU Platforms}
\label{Deploying Testing}
\hl{To evaluate the computational requirements and deployment feasibility of MonitorVLM, we conducted inference experiments on eight NVIDIA H100 GPUs and eight NVIDIA A40 GPUs. Both configurations employed eight-way tensor parallelism, KV caching, a maximum GPU memory utilization setting of 80\%, and a combined input and output token limit of 8,192. During A40 deployment, GPU memory consumption was approximately 36--39 GB per GPU. As shown in Table~}
\ref{tab:gpu_time_comparison}
\hl{, single-request inference latency was 17.59 s on H100 GPUs and 75.70 s on A40 GPUs. At a concurrency level of 60, batch processing times reached 23.45 s and 100.96 s, corresponding to throughputs of 2.56 and 0.59 requests/s, respectively. These results demonstrate that concurrent processing improves overall throughput. Despite their higher inference latency, A40 GPUs offer a more affordable hardware alternative to H100 GPUs for offline safety inspection. Further optimization is required to reduce computational costs and support real-time or edge-based deployment.}
\begin{table*}[!htbp]
\centering
\caption{\hl{Inference Time Comparison Between H100 and A40 GPUs.}}
\label{tab:gpu_time_comparison}
\setlength{\tabcolsep}{12pt}
\begin{tabular}{ccc}
\toprule
\textbf{Concurrency} &
\shortstack{\textbf{H100}\\\textbf{Batch Time (s)}} &
\shortstack{\textbf{A40}\\\textbf{Batch Time (s)}} \\
\midrule
1  & $17.59 \pm 2.11$ & $75.70 \pm 7.57$ \\
10 & $19.54 \pm 2.35$ & $84.16 \pm 8.42$ \\
30 & $21.45 \pm 2.58$ & $92.42 \pm 9.24$ \\
60 & $23.45 \pm 2.82$ & $100.96 \pm 10.10$ \\
\bottomrule
\end{tabular}
\end{table*}

\subsection{\hl{Failure Case Analysis}}
\label{Analysis}
\hl{To better characterize the model's performance and failure modes, we conduct an error analysis on the testing set. In practical mining safety inspections, both confirmed and suspected violations are flagged for manual review. For a violating sample, a prediction is counted as correct only when the model reports a violation and identifies the
annotated violation clause. Fig.~}
\ref{fig.testing_set}
\hl{ shows the recognition counts by violation rule. MonitorVLM-72B achieves a sample-level accuracy of 88.18\% across 2,055 test samples. Fig.~}
\ref{fig.failurecase1} \hl{ summarizes the correct and incorrect outcomes by ground-truth label, including 1,274 true positives (TP, 62.0\%), 538 true negatives (TN, 26.2\%), 148 false negatives (FN, 7.2\%), and 95 false positives (FP, 4.6\%). The false negatives include both missed violations and
violations reported under an incorrect clause.}

\begin{figure}[!ht]
\centering
    \subfigure[Per-rule Recognition Counts]{
    \includegraphics[width=0.96\columnwidth]{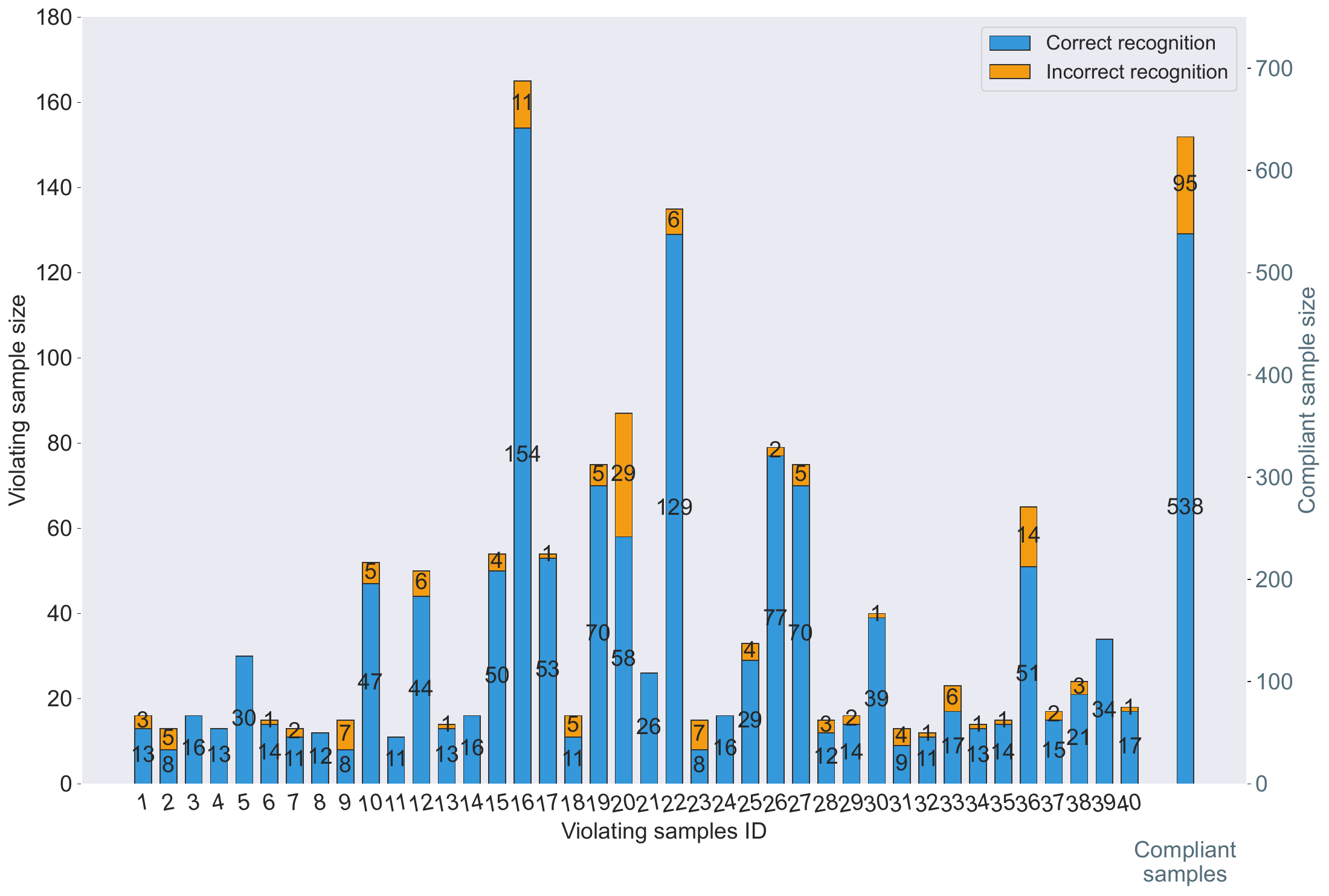}
    \label{fig.testing_set}}
\subfigure[Sample-level evaluation outcomes on the testing set]{
    \includegraphics[width=0.85\columnwidth]{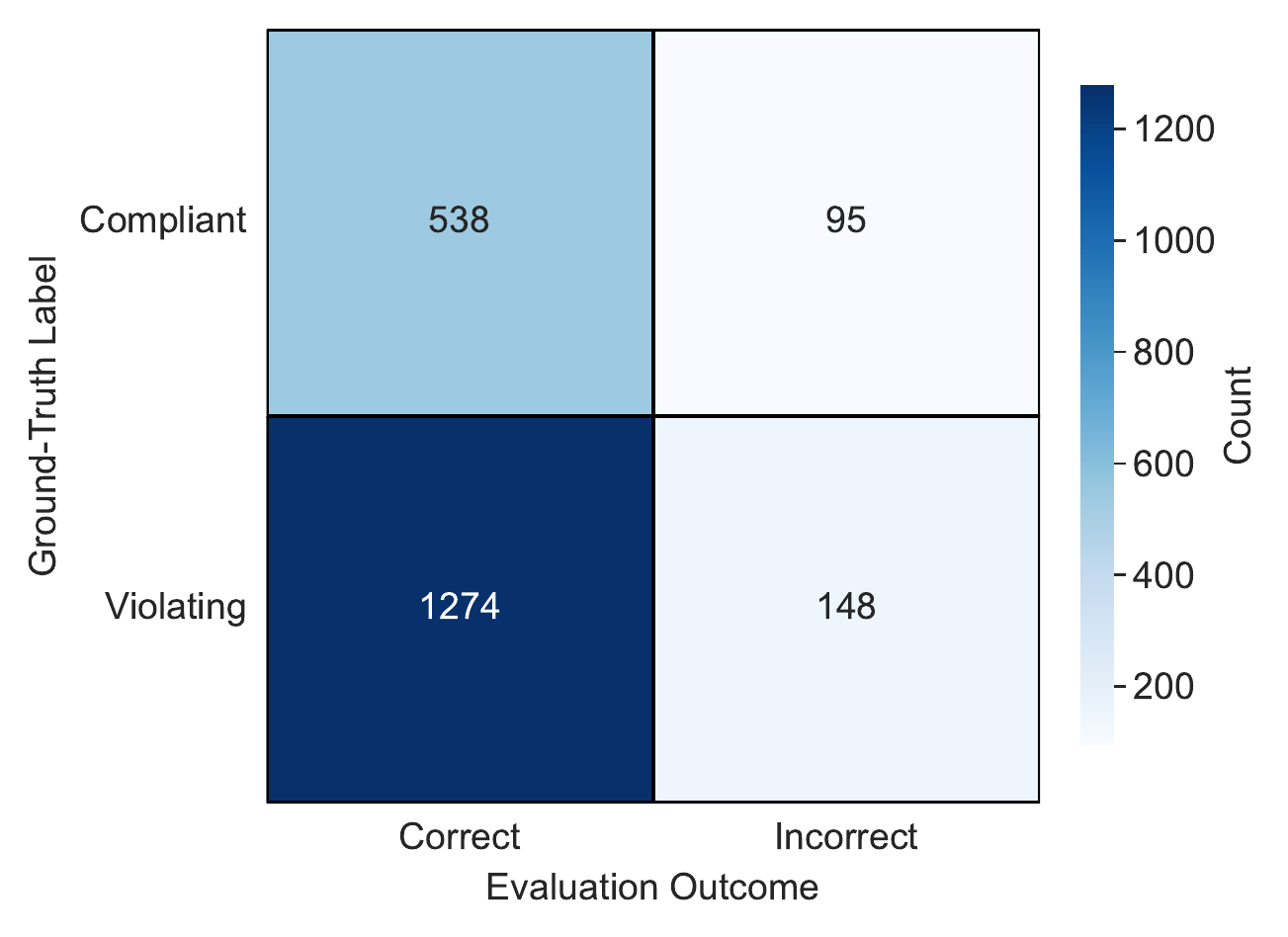}
    \label{fig.failurecase1}}
\caption{{\color{black}Test-set recognition results for MonitorVLM-72B.}}
\end{figure}

To better understand the failure causes, we stratify the 243 error cases (95 FP + 148 FN) by environmental factors. Table~\ref{different violation} summarizes the distribution of errors and the associated failure modes across four challenging conditions, reporting the proportion of each error type under different settings.

\begin{table*}[!h]
\centering
\caption{{\color{black}Error Distribution and Failure Modes under Challenging Environmental Conditions.}}
\label{different violation}
\begin{tabular}{lcp{6cm}}
\toprule
\textbf{Environmental Factor} & \textbf{Error Proportion} & \textbf{Primary Failure Modes} \\
\midrule   
Low/High Lighting & 42.80\% & 
    \textbf{FN:} Missed small objects (helmets, safety belts) in shadows or glare \newline
    \textbf{FP:} Light sources (headlamps, screens) misidentified as electronic devices \\
\midrule
Heavy Occlusion & 23.46\% & 
    \textbf{FN:} Incomplete visual evidence due to machinery/equipment blocking \newline
    \textbf{FP:} Partial observations misinterpreted as violations \\
\midrule
Camera Viewpoint & 25.10\% & 
    \textbf{FN:} Fine-grained actions undetectable from distant/oblique angles \newline
    \textbf{FP:} Ambiguous PPE status (e.g., chin strap unclear) \\
\midrule
Multiple Overlapping Workers & 8.64\% & 
    \textbf{FN:} Inter-person occlusion and inconsistent behaviors across workers \newline
    \textbf{FP:} Token-limit constraints causing incomplete scene understanding \\
\bottomrule
\end{tabular}
\end{table*}

Fig.~\ref{Environmental factors.} provides visual examples of these failure-prone scenarios (Fig.~\ref{fig.case1}--\ref{fig.case5}). Specifically, extreme lighting conditions can cause the model to miss small objects (e.g., safety belts) and may trigger false detections of electronic device screens due to specular reflections. Heavy occlusion often results in insufficient visual evidence for confirming violations, or leads to misinterpretation of partially visible workers. Distant camera viewpoints hinder fine-grained action recognition and make PPE status ambiguous. Finally, scenes with multiple overlapping workers may introduce token-limit constraints and lead to inconsistent behavior predictions across individuals. \hl{To address these limitations, future improvements could include optimizing camera placement and viewing angles in collaboration with on-site surveillance personnel to reduce occlusion and improve visibility of distant workers. For images affected by poor lighting or low visibility, lightweight image enhancement models could be introduced as a preprocessing stage to perform adaptive denoising and illumination correction before VLM inference. Furthermore, extending temporal modeling over longer video sequences could help distinguish individual worker behaviors and improve detection consistency in crowded scenes.}

\begin{figure}[!htbp]
\centering
\subfigure[High lighting]{
    \includegraphics[width=0.45\columnwidth]{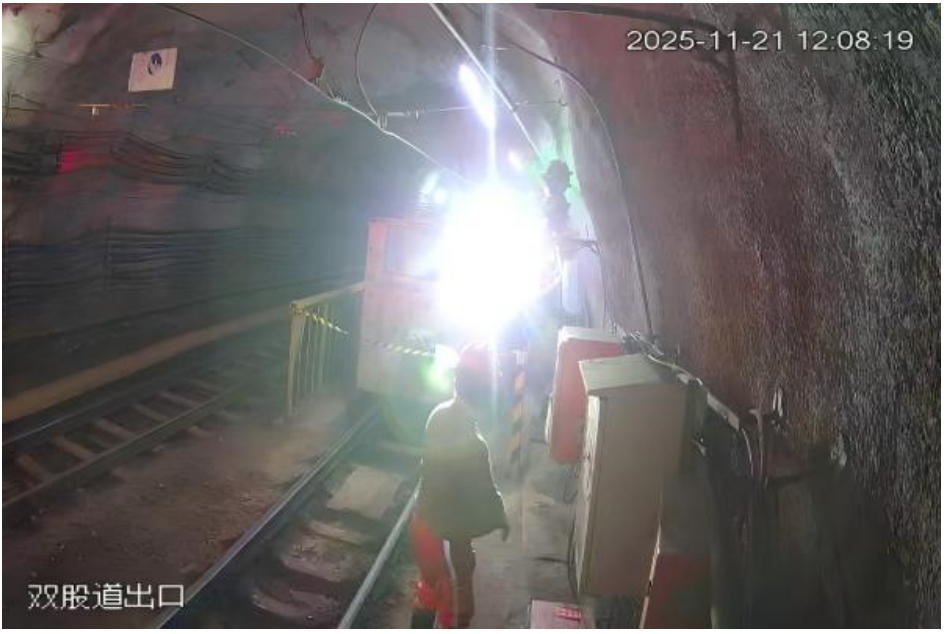}
    \label{fig.case1}}
\subfigure[Low lighting]{
    \includegraphics[width=0.45\columnwidth]{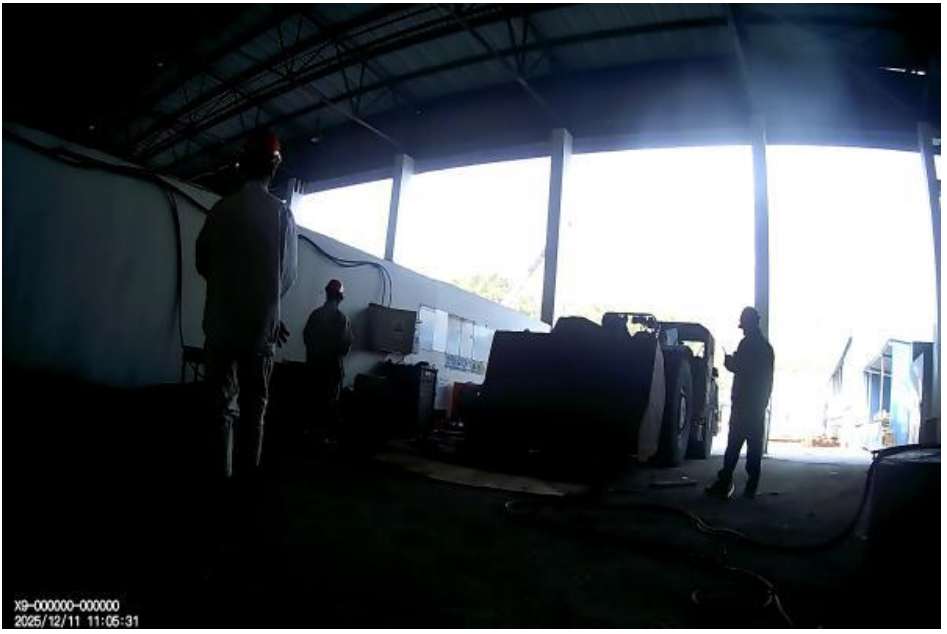}
    \label{fig.case2}}
\subfigure[Heavy occlusion]{
    \includegraphics[width=0.45\columnwidth]{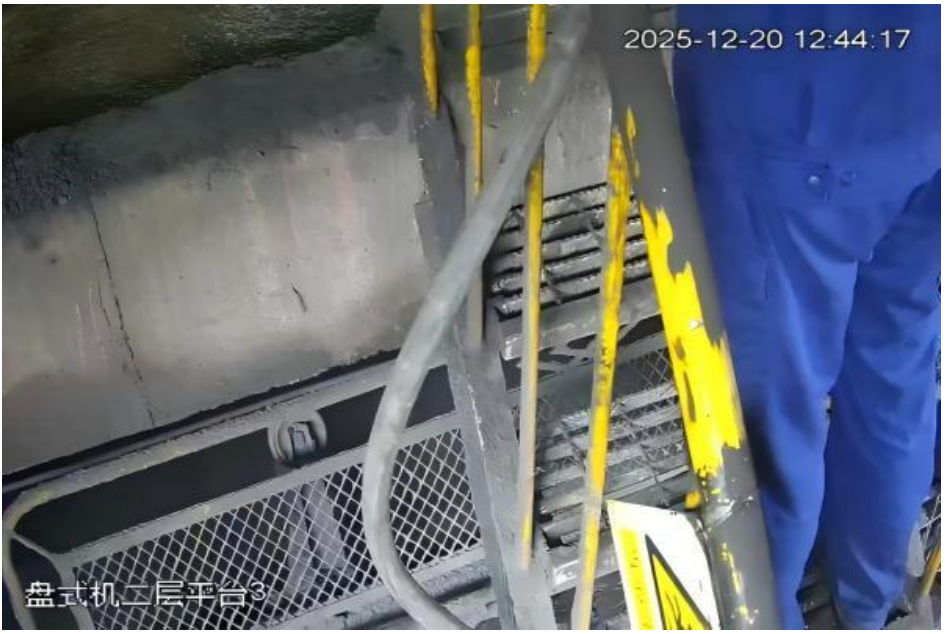}
    \label{fig.case3}}
\subfigure[Camera viewpoint]{
    \includegraphics[width=0.45\columnwidth]{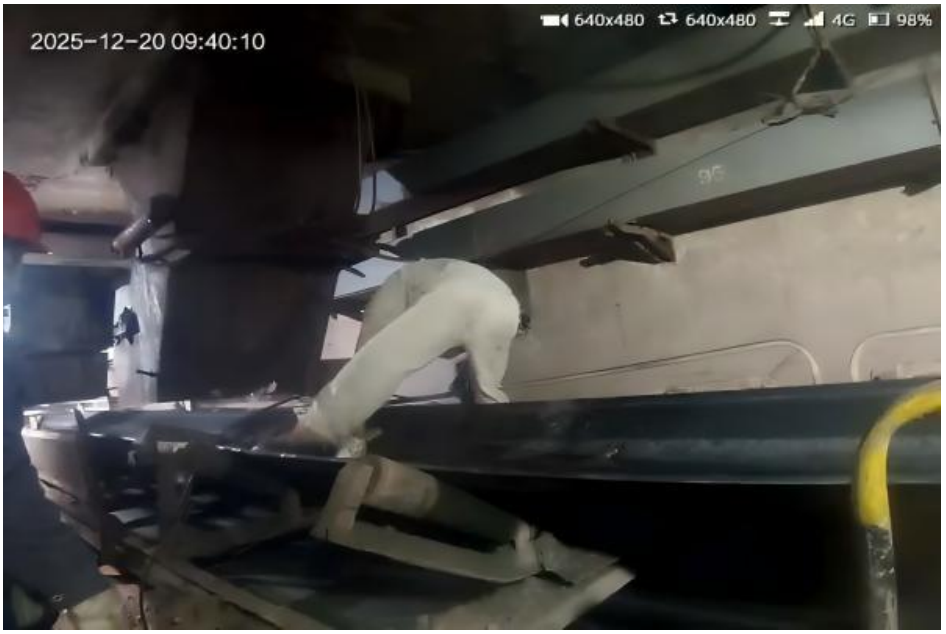}
    \label{fig.case4}}
    \subfigure[Multiple overlapping workers]{
    \includegraphics[width=0.45\columnwidth]{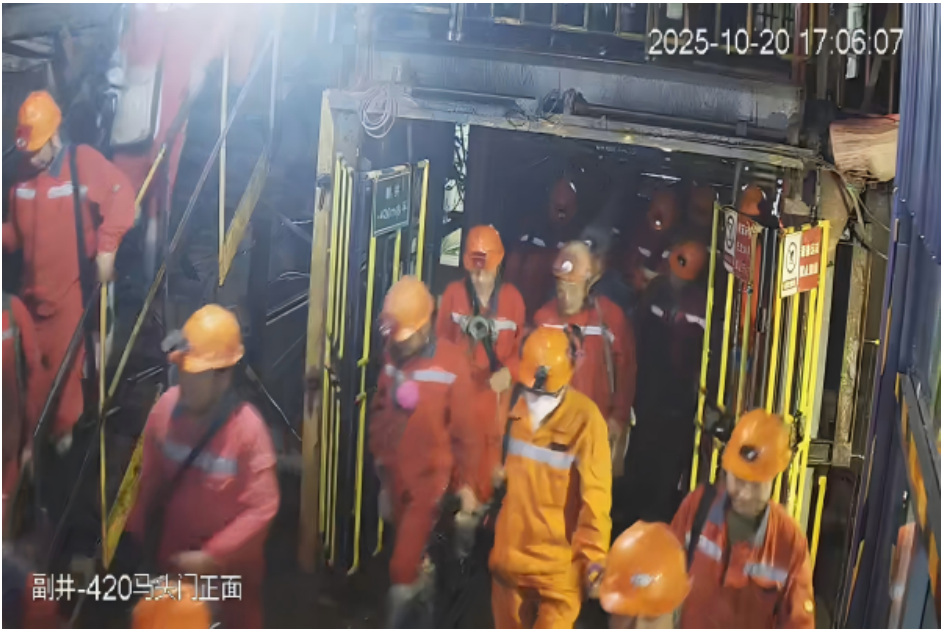}
    \label{fig.case5}}
\caption{{\color{black}Failure-prone scenarios.}}
\label{Environmental factors.}
\end{figure}

}

\section{Conclusion}
\label{Conclusion}
\hl{This work presented MonitorVLM, a vision--language framework for automated safety violation detection in offline mining surveillance videos. The experimental findings demonstrate that integrating domain-specific knowledge with vision--language reasoning can effectively improve the identification of unsafe worker behaviors and their association with relevant safety regulations. In particular, domain-specific fine-tuning strengthens the model's understanding of mining scenarios, while the dual-encoder clause filter reduces unnecessary regulatory information without compromising detection performance. The behavior magnifier further improves the recognition of subtle worker actions under challenging visual conditions. These findings highlight the importance of combining task-specific adaptation, efficient regulatory knowledge selection, and fine-grained visual understanding for reliable safety inspection. Overall, MonitorVLM provides an effective approach to rule-oriented safety analysis, offering a practical foundation for more automated and interpretable safety monitoring in mining operations.}

\hl{Although this paper has achieved some progress, several limitations
remain that warrant further exploration and improvement in future
work. The following directions are proposed for future research:}

1) \hl{Dataset Scale and Diversity: Our dataset covers 40 high-frequency mining  regulations and surveillance footage from more than 77 cameras within a single mining site. However, cross-site generalization remains unverified. Future work will focus on collecting data from additional mining sites and conducting cross-site validation to further evaluate the generalizability of MonitorVLM.}

2)\hl{Computational Efficiency: Although MonitorVLM is currently designed for offline video inspection, real-time monitoring is essential for the timely detection of unsafe behaviors and prevention of potential accidents. Future work will focus on reducing inference latency and computational requirements to enable real-time safety monitoring in mining operations.}

\section*{Appendix}
\label{Appendix}
\begin{figure}[!htbp]
\centering
\subfigure[Loading Warehouse]{
    \includegraphics[width=0.38\columnwidth]{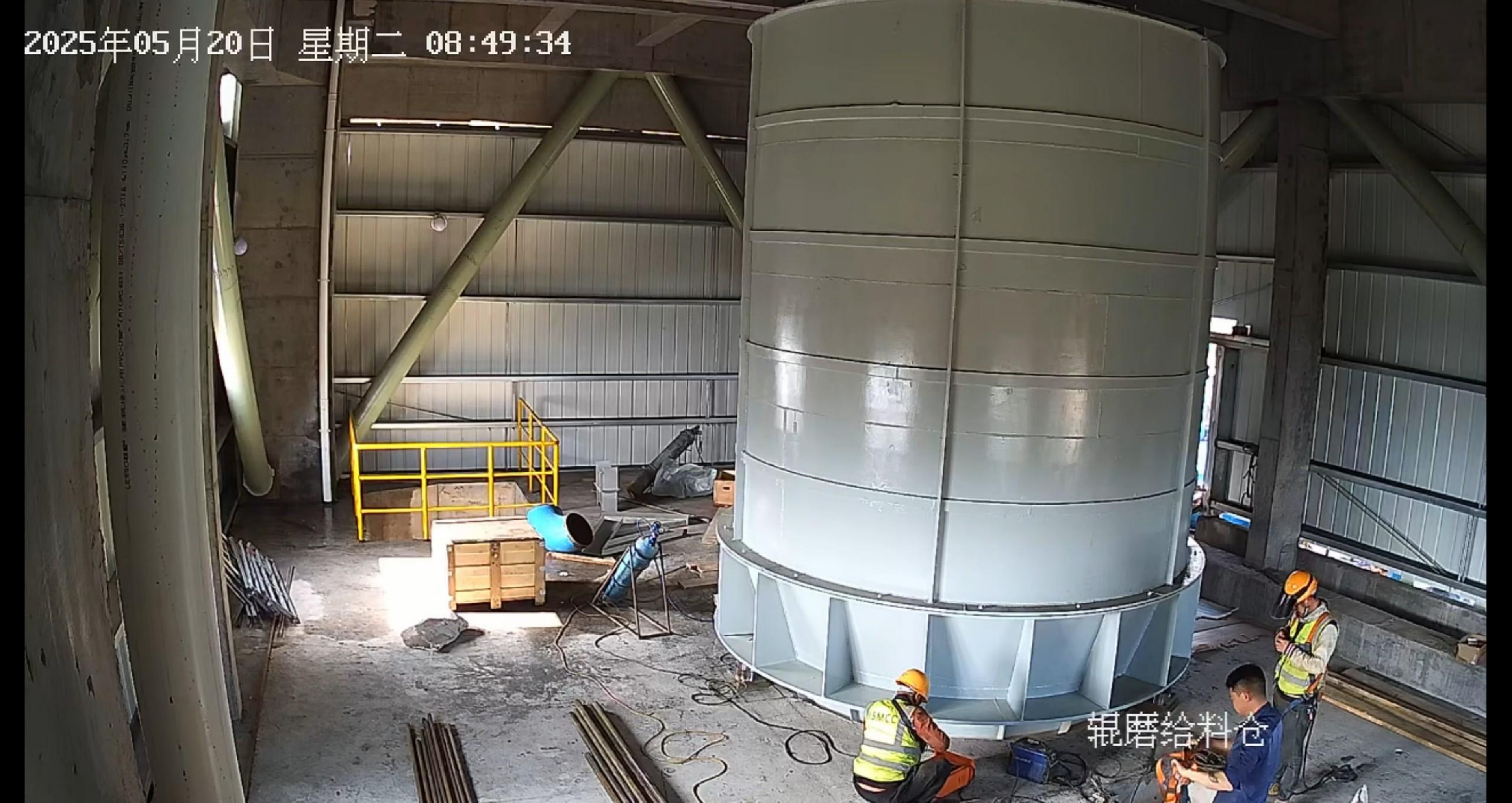}
   }
\subfigure[Unloading warehouse]{
    \includegraphics[width=0.38\columnwidth]{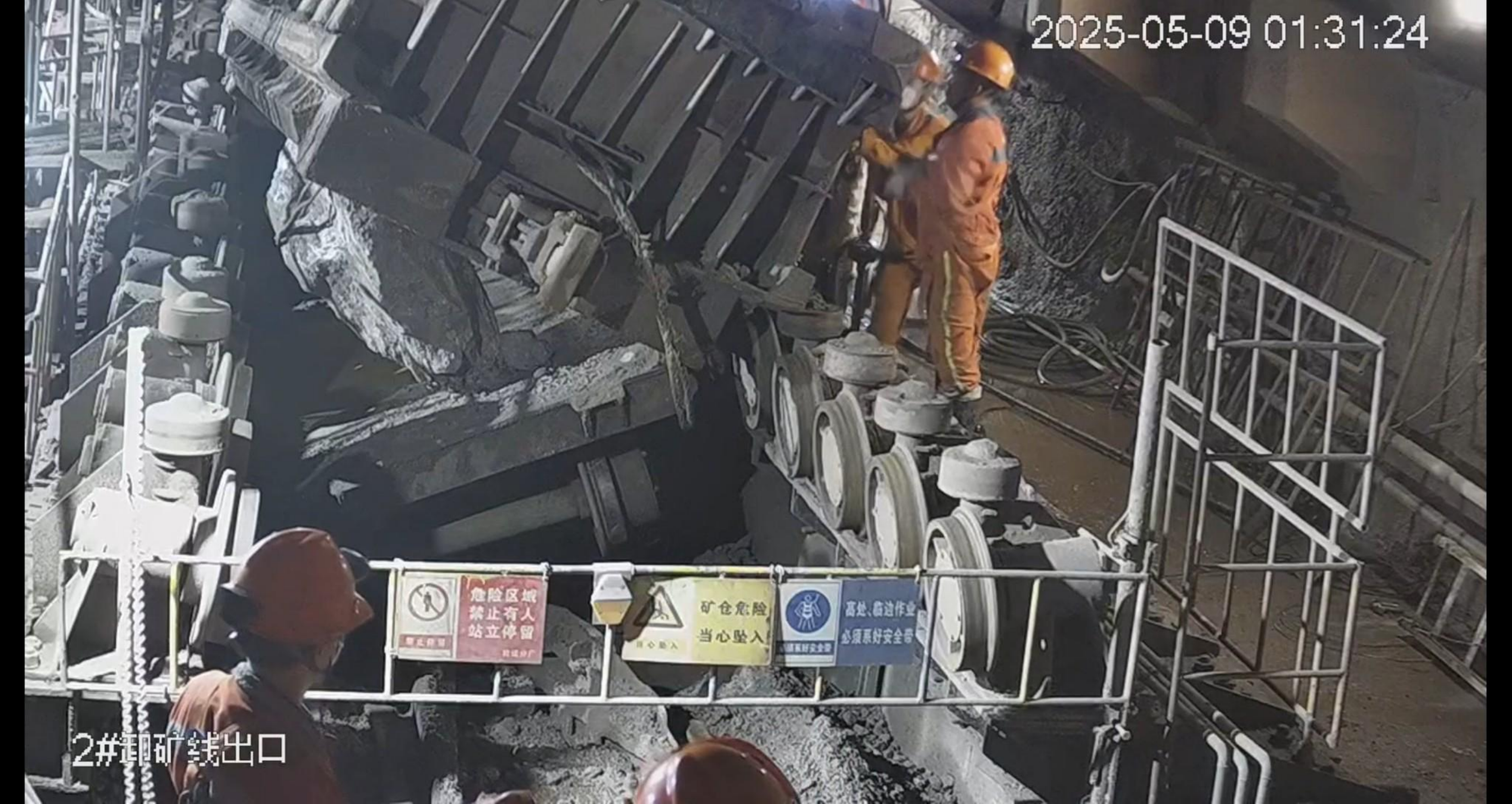}
  }
\subfigure[Aisle escalator]{
\includegraphics[width=0.38\columnwidth]{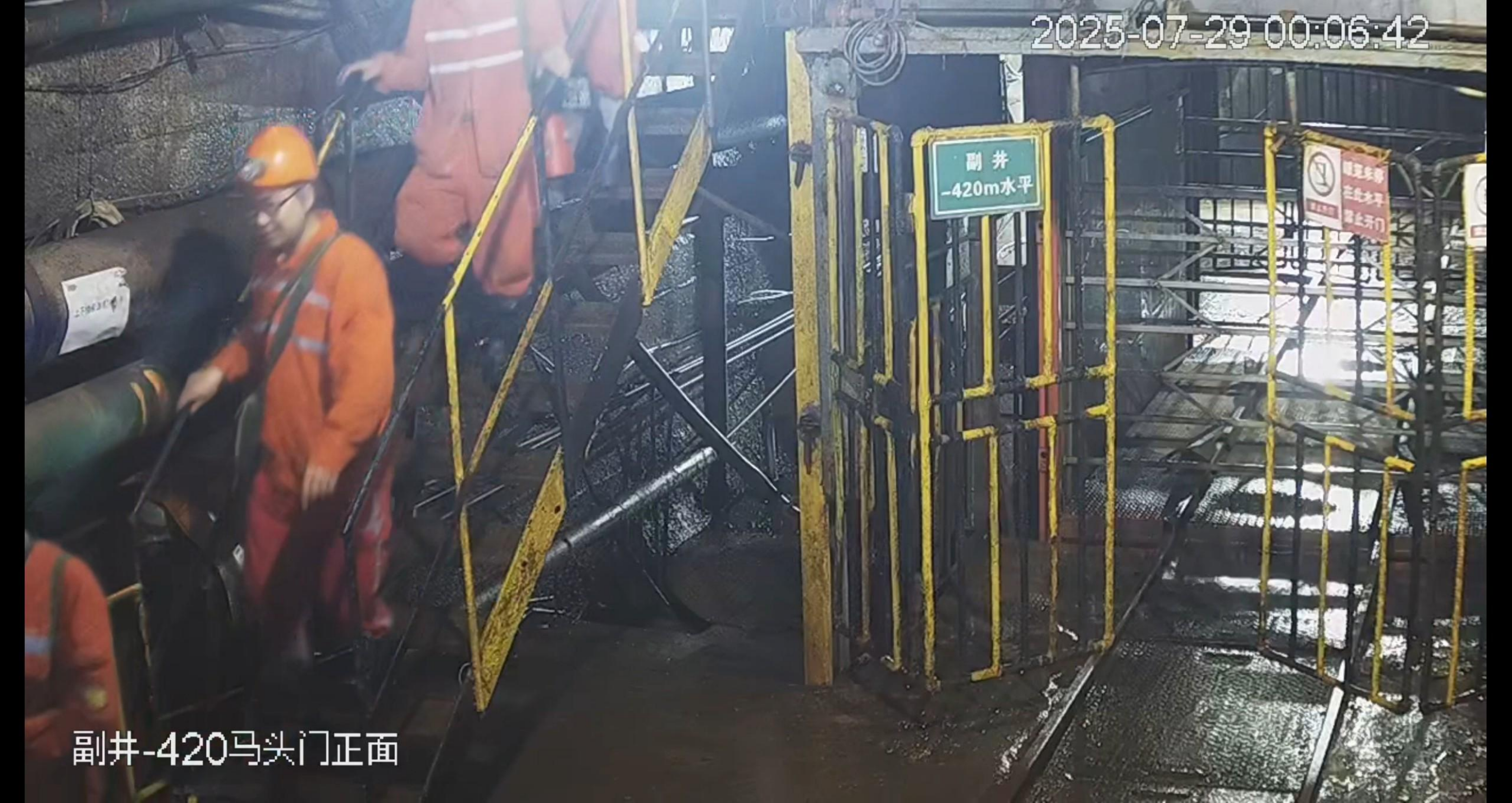}
    }
    \subfigure[Welding area in the tunnel]{
\includegraphics[width=0.38\columnwidth]{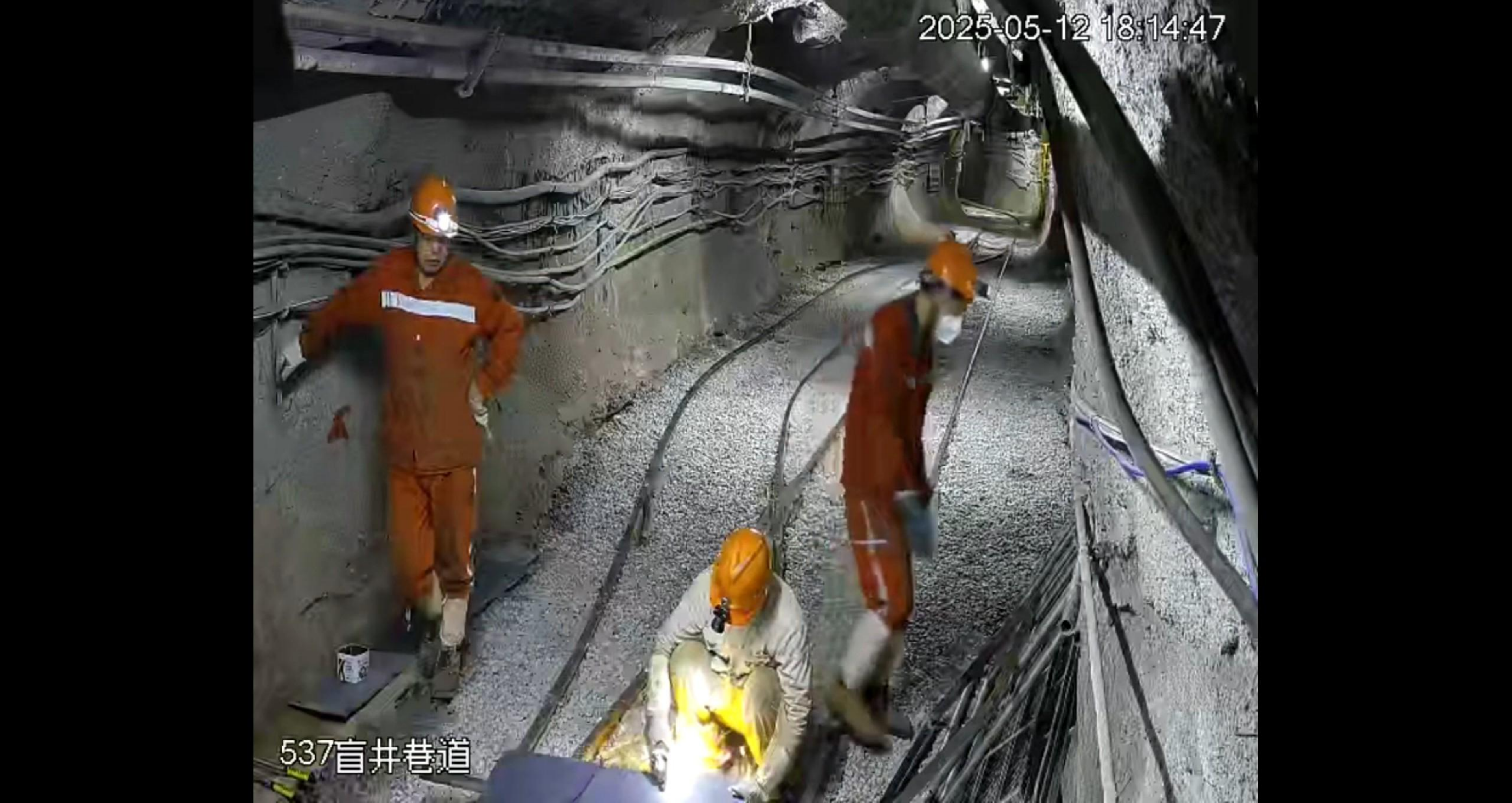}
    }
    \subfigure[Area near the conveyor belt]{
\includegraphics[width=0.38\columnwidth]{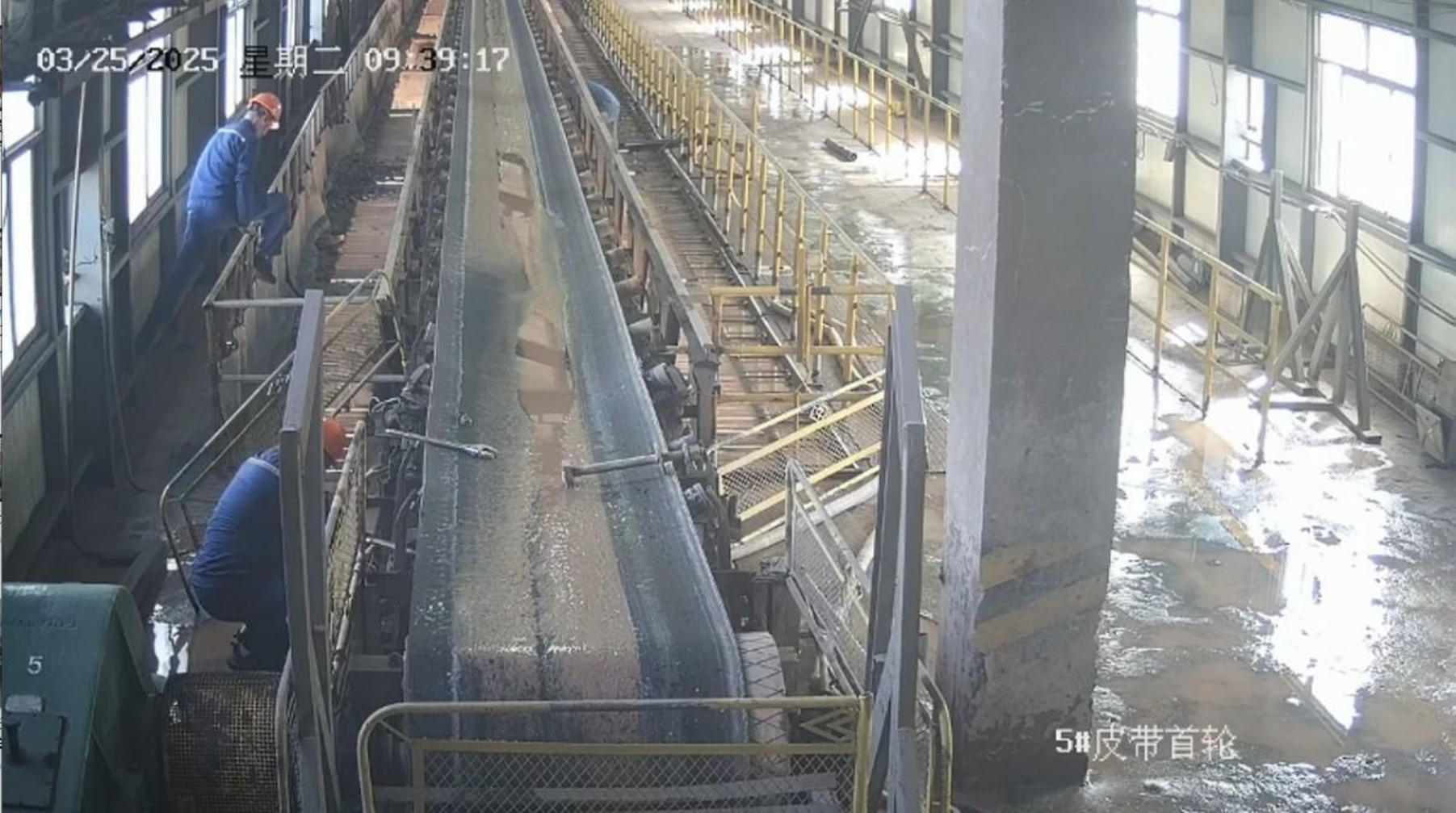}
    }
    \subfigure[Crane operation zone]{
\includegraphics[width=0.38\columnwidth]{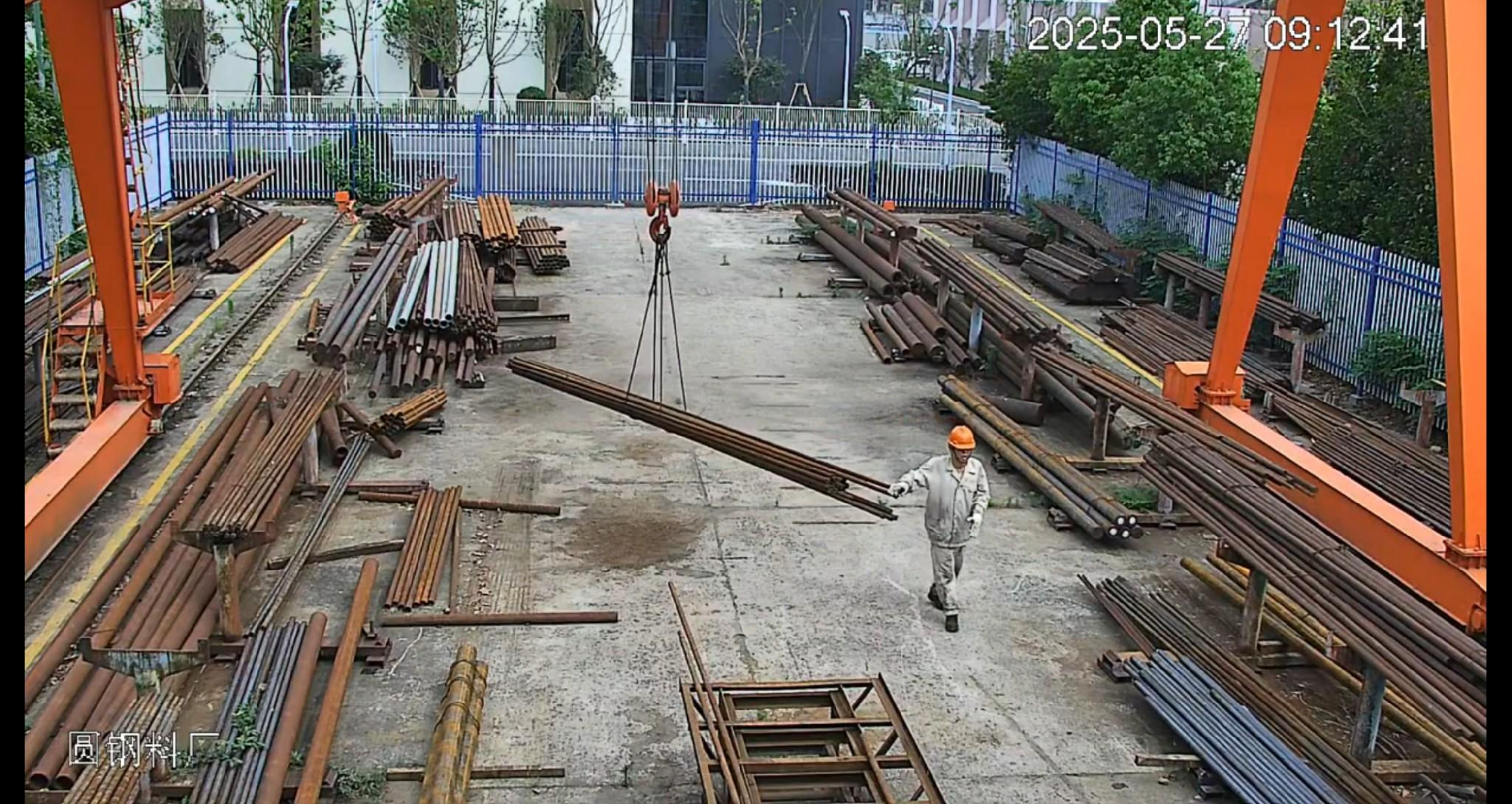}
    }
    \subfigure[Machining and repair area]{
\includegraphics[width=0.38\columnwidth]{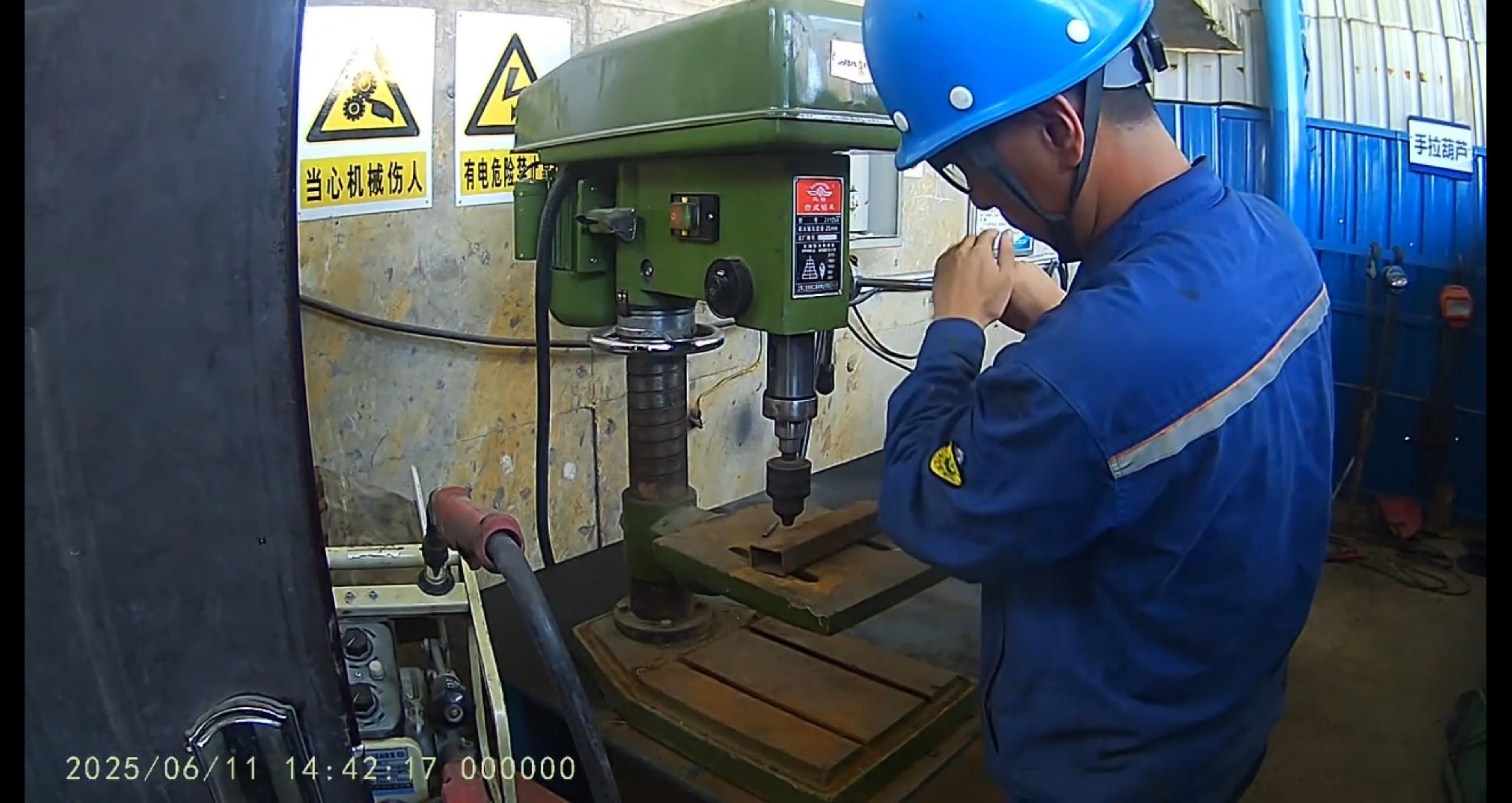}
    }
        \subfigure[Discharging area]{
\includegraphics[width=0.38\columnwidth]{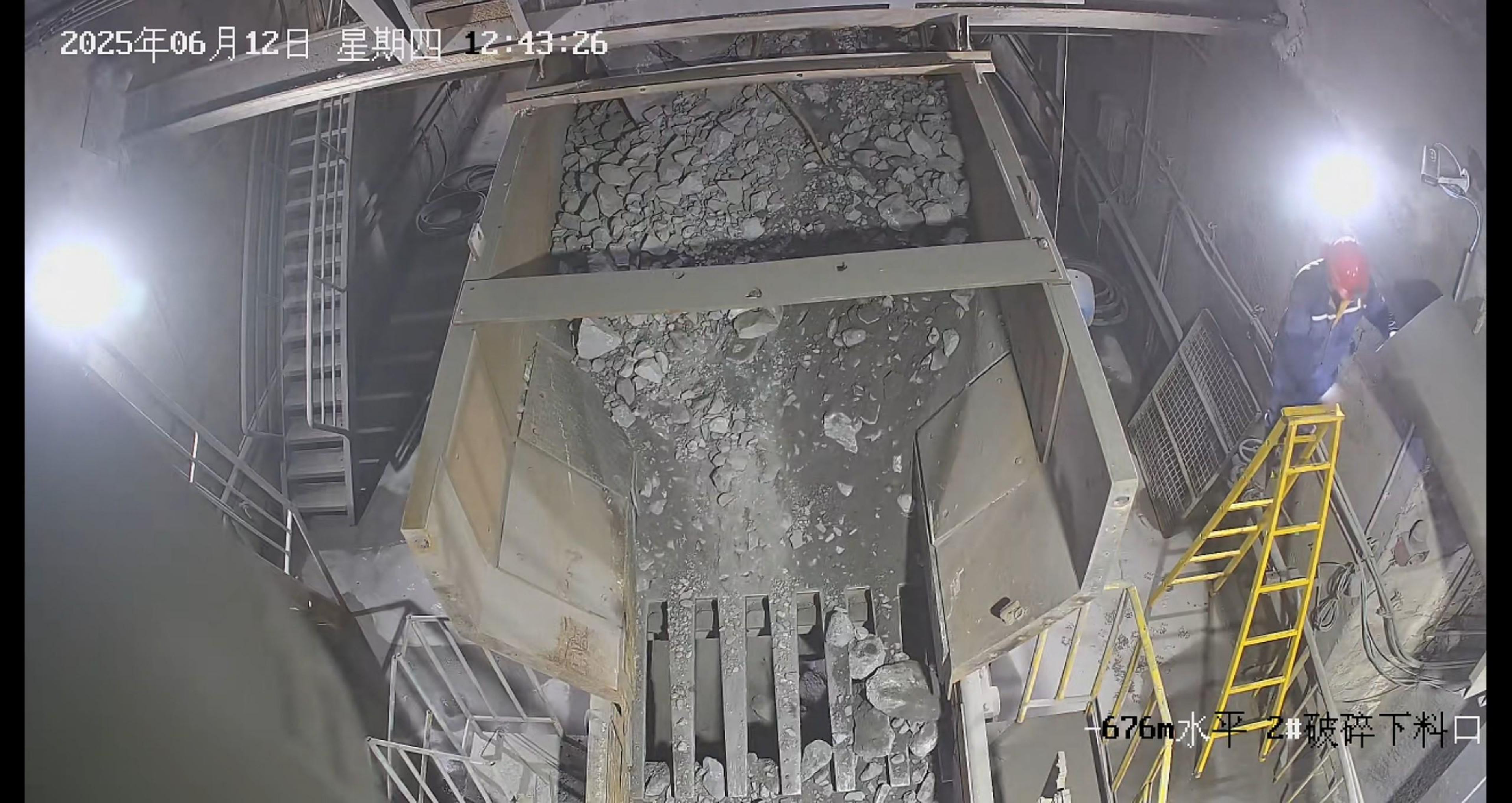}
    }
        \subfigure[Control room]{
\includegraphics[width=0.38\columnwidth]{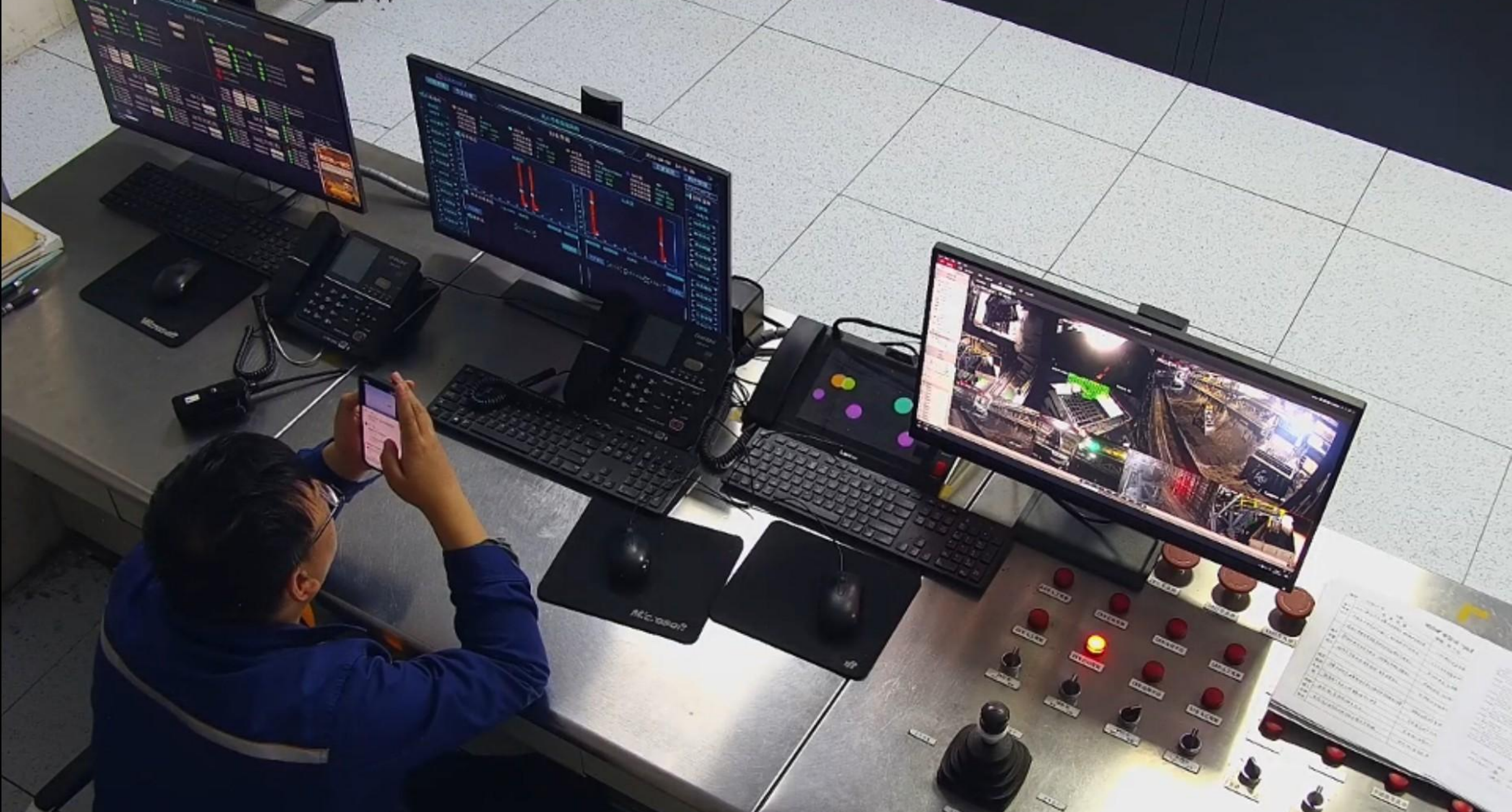}
    }
            \subfigure[Tunnel machine working area]{
\includegraphics[width=0.38\columnwidth]{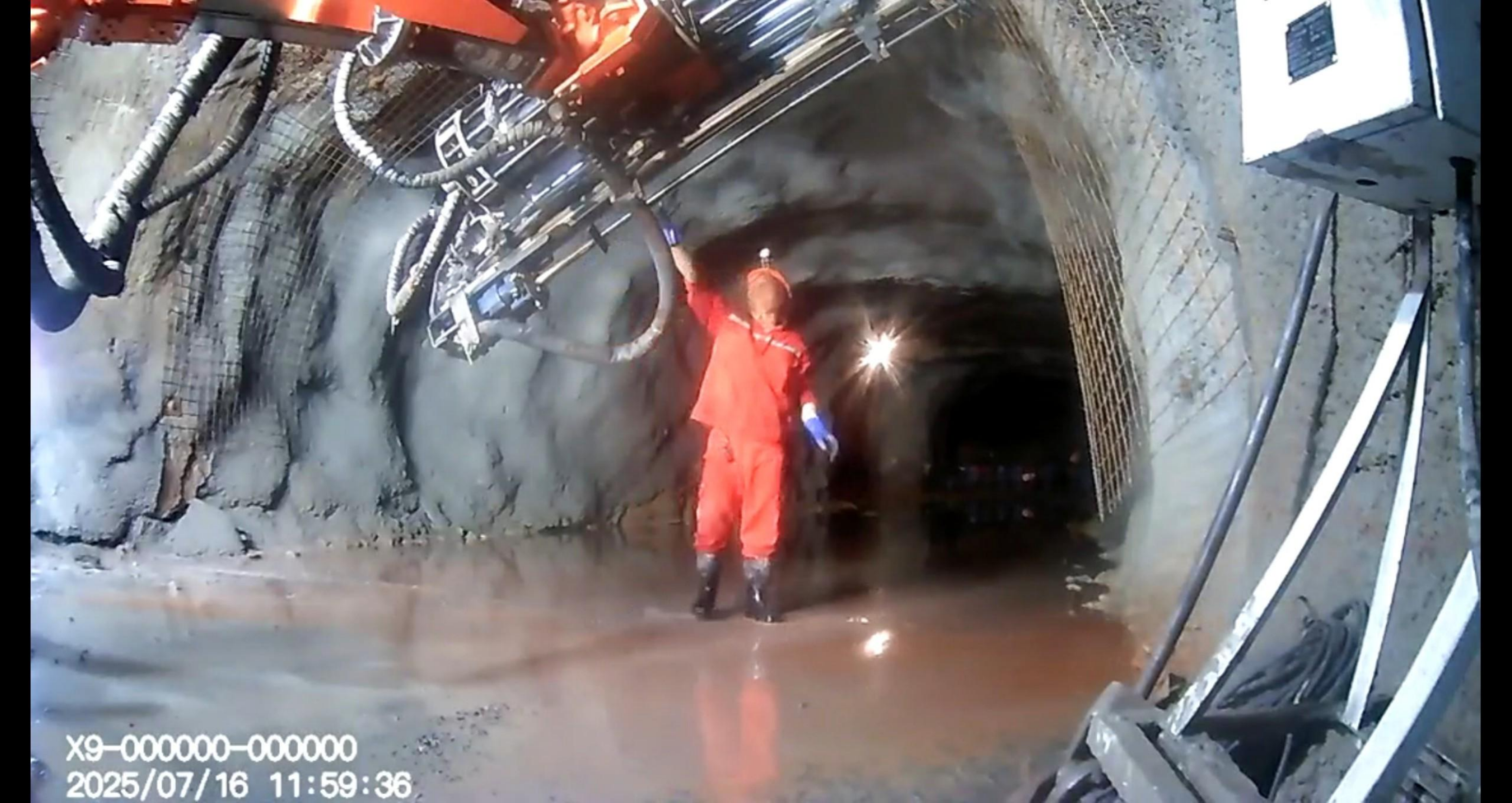}
    }

\caption{{\color{black}Representative surveillance images of 10 operational areas.}}
\label{10areas}
\end{figure}

\begin{figure}[!htbp]
  \centering
  \includegraphics[width=0.96\columnwidth]{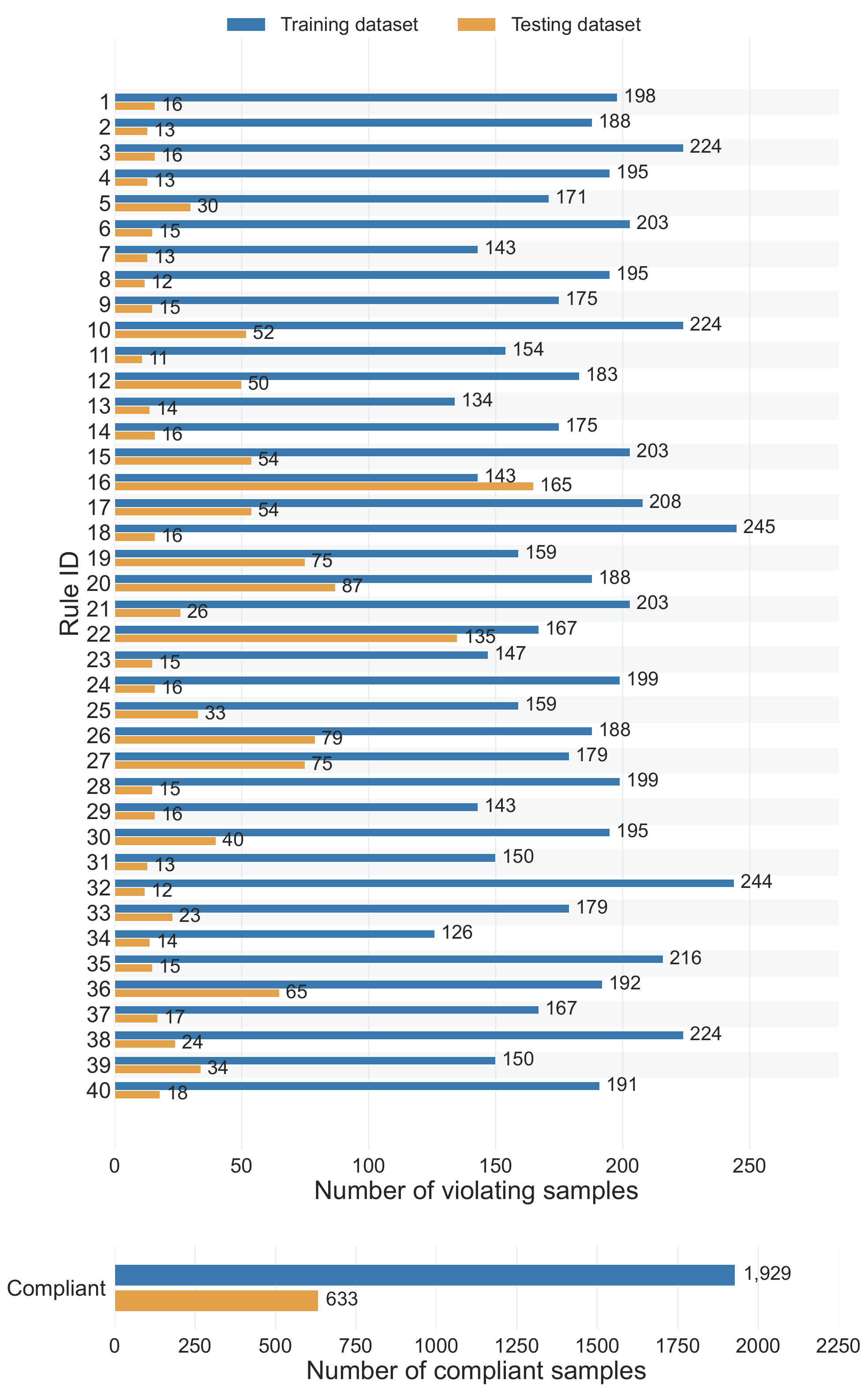}
  \caption{{\hl{The distribution of training and testing samples.}}}
  \label{categories}
\end{figure}

\hl{This Appendix presents 10 representative operational areas in Fig.~}
\ref{10areas}
\hl{, including loading warehouses, unloading warehouses, passageway escalators, tunnel welding areas, conveyor belt areas, crane operation zones, machining and maintenance areas, unloading zones, control rooms, and tunnel machinery operation areas. The dataset covers 23 operational areas within a single mining site, with surveillance footage collected from more than 77 cameras. Since several areas share similar operational characteristics, only 10 representative areas are illustrated. Recordings span different times throughout the day, with an emphasis on periods of active worker operations. The 40 high-frequency safety regulations are listed in Table~}
\ref{tab:unsafe_practices}
\hl{ and grouped into three categories: unsafe worker behaviors, unsafe use of tools and equipment, and personal protective equipment (PPE) issues. Fig.~}
\ref{categories}
\hl{ summarizes the sample distributions. The training dataset contains 9,255 samples, including 7,326 violation samples (79.16\%) and 1,929 compliant samples (20.84\%). The testing dataset comprises 2,055 samples, including 1,422 violation samples (69.20\%) and 633 compliant samples (30.80\%).}

\begin{table*}[hbtp]
\centering
\caption{List of 40 high-frequency and high-risk mining violations, organized by category: unsafe worker behavior, unsafe use of tools and equipment, and PPE.}
\label{tab:unsafe_practices}
\begin{tabularx}{\textwidth}{c p{4.8cm} X}
\toprule
\textbf{No.} & \textbf{Category} & \textbf{Unsafe Practice} \\
\midrule
1 & Unsafe worker behavior & Extending any part of the body outside a moving personnel carrier or trackless vehicle \\
2 & Unsafe worker behavior & Standing in front of a grinder or cutting machine during cutting operations \\
3 & Unsafe worker behavior & Removing metal chips directly by hand \\
4 & Unsafe use of tools and equipment & Transporting excessively wide loads with a forklift without protective measures \\
5 & Unsafe worker behavior & Removing metal chips by hand or with a cloth while a drill press is running \\
6 & Unsafe use of tools and equipment & Operating underground equipment or vehicles without lighting \\
7 & Unsafe worker behavior & Carrying passengers on a forklift \\
8 & Unsafe worker behavior & Passing objects across machinery while it is operating \\
9 & Unsafe worker behavior & Standing within the projected fall area of a large object while it is being overturned \\
10 & PPE & Climbing without using a double-hook safety harness or fall arrester \\
11 & Unsafe worker behavior & Carrying more than one person in the driver's position of an electric tricycle \\
12 & PPE & Using an angle grinder without protective goggles \\
13 & Unsafe use of tools and equipment & Parking a load-haul-dump vehicle or loader without lowering its bucket to the ground \\
14 & Unsafe worker behavior & Throwing objects from height \\
15 & Unsafe worker behavior & Standing within the swept area of an equipment boom while it is moving \\
16 & PPE & Not wearing a safety helmet in a production work area \\
17 & PPE & Failing to secure long hair inside a safety helmet during work \\
18 & Unsafe worker behavior & Climbing railings on a platform with a fall height of 1--2 m without protective measures \\
19 & Unsafe worker behavior & Using a mobile phone in a work area \\
20 & Unsafe use of tools and equipment & Working near a belt conveyor without stopping it \\
21 & Unsafe worker behavior & Boarding or alighting from lifting machinery during startup \\
22 & PPE & Working near an unprotected edge or opening without a safety harness \\
23 & PPE & Working in a specialized laboratory without protective footwear \\
24 & Unsafe worker behavior & Smoking in the workplace \\
25 & PPE & Personnel near cutting operations failing to wear masks \\
26 & PPE & Wearing or attaching a safety harness improperly, including loose fastening \\
27 & Unsafe use of tools and equipment & Operating a material or personnel transport vehicle with its doors open \\
28 & Unsafe use of tools and equipment & Starting equipment without authorization after it has been shut down and tagged \\
29 & Unsafe use of tools and equipment & Placing objects inside or on top of electrical distribution boxes or cabinets \\
30 & PPE & Failing to fasten a seat belt while a vehicle is moving \\
31 & Unsafe worker behavior & Rolling oxygen or acetylene cylinders along the ground during transport \\
32 & Unsafe use of tools and equipment & Driving into and damaging railings around an ore pass \\
33 & Unsafe use of tools and equipment & Performing maintenance or construction work while standing inside a raised loader bucket \\
34 & Unsafe worker behavior & Standing in front of a mine car to pull it \\
35 & Unsafe worker behavior & Using stairs without holding the handrail \\
36 & PPE & Using a double-hook safety harness or fall arrester incorrectly while climbing \\
37 & Unsafe worker behavior & Jumping down from a height of 50 cm \\
38 & Unsafe worker behavior & Standing too close to lifting machinery during startup \\
39 & PPE & Performing cutting operations without the required face protection or gloves \\
40 & Unsafe use of tools and equipment & Using a raised loader bucket for access to height or for tasks other than maintenance or construction \\
\bottomrule
\end{tabularx}
\end{table*}

\bibliographystyle{IEEEtran}
\bibliography{reference}
\end{document}